%% file: main.tex
\documentclass[10pt,twocolumn,letterpaper]{article}

\usepackage[pagenumbers]{cvpr} 

\input{preamble}

\usepackage{fancyvrb}
\usepackage[most]{tcolorbox}
\usepackage{listings}
\lstdefinestyle{mystyle}{
  basicstyle=\ttfamily,
  breaklines=true,
  breakautoindent=true,
  breakindent=2em,
  moredelim=[is][\bfseries]{@}{@}
}
\usepackage[utf8]{inputenc} 
\usepackage[T1]{fontenc}    
\usepackage{url}            
\usepackage{booktabs}       
\usepackage{amsfonts}       
\usepackage{nicefrac}       
\usepackage{microtype}      
\usepackage{xcolor}         
\usepackage{amsmath}  
\usepackage{algorithm}
\usepackage{multirow}
\usepackage{algorithmicx}
\usepackage{algpseudocode}
\usepackage{booktabs}
\usepackage{makecell}
\usepackage{bbding}
\usepackage{pifont}
\usepackage{xcolor}
\usepackage{colortbl}
\usepackage{graphicx}
\usepackage{enumitem}
\usepackage{caption}
\usepackage{amssymb}   

\DeclareMathOperator*{\argmax}{arg\,max}

\definecolor{cvprblue}{rgb}{0.21,0.49,0.74}
\usepackage[pagebackref,breaklinks,colorlinks,allcolors=cvprblue]{hyperref}

\def\paperID{} 
\def\confName{CVPR}
\def\confYear{2026}

\title{Learning Mutual View Information Graph for Adaptive Adversarial Collaborative Perception}

\author{Yihang Tao$^{1,2}$, Senkang Hu$^{1,2}$, Haonan An$^{1,2}$, Zhengru Fang$^{1,2}$,  Hangcheng Cao$^{1,2}$, Yuguang Fang$^{1,2}$  \\ 
$^1$Hong Kong JC STEM Lab of Smart City,
$^2$City University of Hong Kong,\\
{\tt \footnotesize \{yihang.tommy, senkang.forest, haonanan2-c, zhefang4-c\}@my.cityu.edu.hk,}\\
{\tt \footnotesize \{hangccao, my.Fang\}@cityu.edu.hk} 
\vspace{-8mm}
}

\begin{document}
\maketitle

\begin{abstract}
    Collaborative perception (CP) enables data sharing among connected and autonomous vehicles (CAVs) to enhance driving safety. However, CP systems are vulnerable to adversarial attacks where malicious agents forge false objects via feature-level perturbations. Current defensive systems use threshold-based consensus verification by comparing collaborative and ego detection results. 
    Yet, these defenses remain vulnerable to more sophisticated attack strategies that could exploit two critical weaknesses: (i) lack of robustness against attacks with systematic timing and target region optimization, and (ii) inadvertent disclosure of vulnerability knowledge through implicit confidence information in shared collaboration data. 
    In this paper, we propose \texttt{MVIG} attack, a novel adaptive adversarial CP framework learning to capture vulnerability knowledge disclosed by different defensive CP systems from a \textbf{unified mutual view information graph (MVIG) representation}. 
    Our approach combines MVIG representation with temporal graph learning to generate evolving fabrication risk maps and employs entropy-aware vulnerability search to optimize attack location, timing and persistence, enabling adaptive attacks with generalizability across various defensive configurations. Extensive evaluations on OPV2V and Adv-OPV2V datasets demonstrate that \texttt{MVIG} attack reduces defense success rates by up to 62\% against state-of-the-art defenses while achieving 47\% lower detection for persistent attacks at 29.9 FPS, exposing critical security gaps in CP systems.
    Code will be released at \url{https://github.com/yihangtao/MVIG.git}
\end{abstract}

\input{sections/intro.tex}

\input{sections/related_works.tex}
\input{sections/method.tex}

\input{sections/experiments.tex}

\input{sections/conclusion_ackno.tex}

{
    \small
    \bibliographystyle{ieeenat_fullname}
    \bibliography{main}
}

\input{sections/appendix.tex}

\end{document}

%% file: preamble.tex
\usepackage{multirow}

%% file: sections/intro.tex
\section{Introduction}

\begin{figure}[t]
  \centering
  \includegraphics[width=1.0\linewidth]{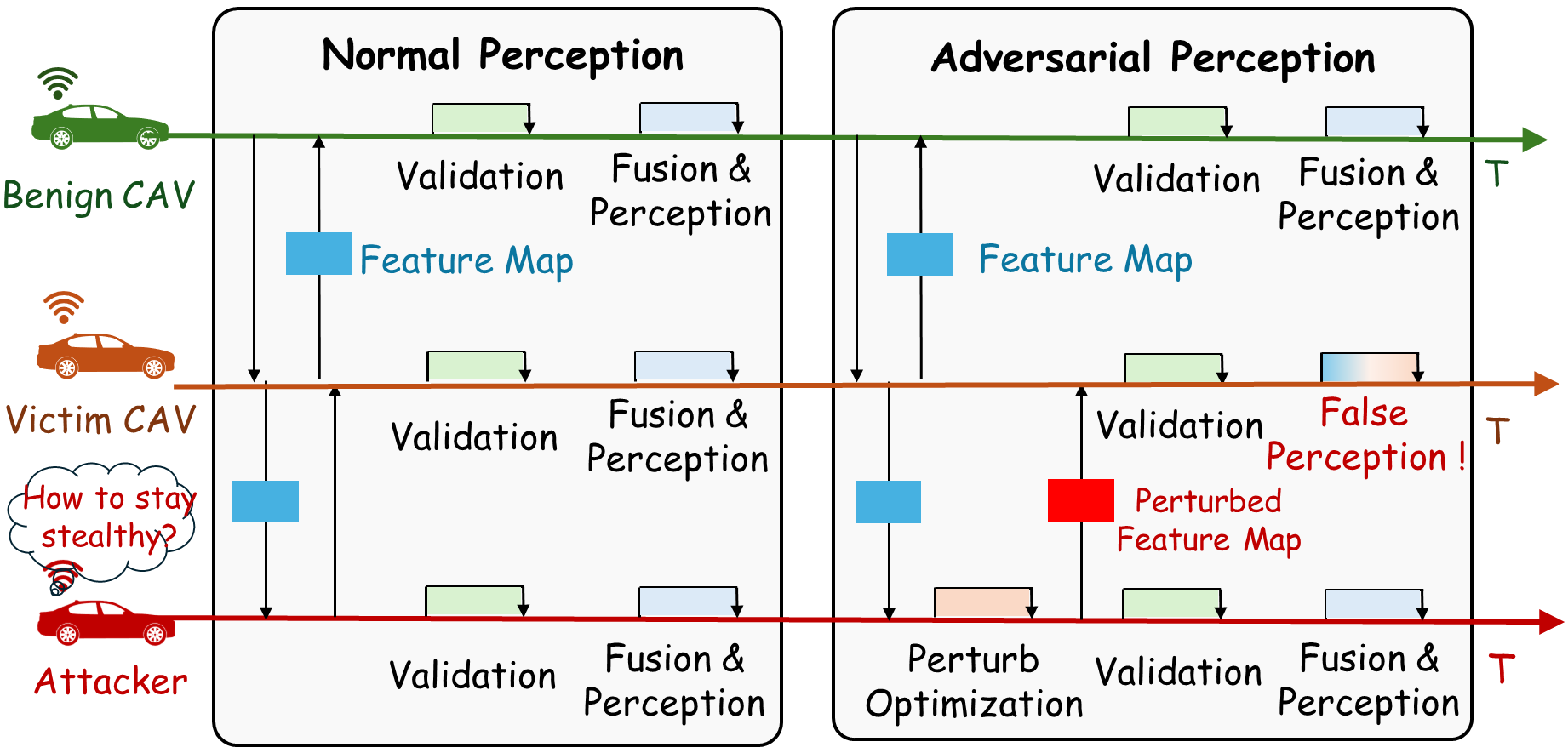}
  \vspace{-5mm}
  \caption{Overview of adversarial attacks targeting CP systems.}
  \label{fig:scenario}
  \vspace{-6mm}
\end{figure}

Recent advances in connected and autonomous vehicles (CAVs) have enabled collaborative perception (CP), allowing vehicles to perceive beyond their immediate field of view (FoV) and address occlusion challenges \cite{chenCooperCooperativePerception2019,liV2XSimMultiAgentCollaborative2022,huAdaptiveCommunicationsCollaborative2023,huWhere2commCommunicationefficientCollaborative2024, tao_directcp}. 
While CP systems are promising, relying on communication between agents poses security issues because the shared information can be malicious or unreliable.
Although modern cryptographic algorithms provide robust protection against external communication breaches, they cannot fully safeguard CP systems from internal threats.
For instance, a CAV's owner with privileged access might intentionally alter their communication data (e.g.,  feature maps) to manipulate traffic flow for personal gain. Besides, skilled attackers could infiltrate legitimate CAVs, gaining unauthorized access to software or hardware, and transmit deceptive messages to disrupt traffic or cause collisions.
These internal attack scenarios in CP have been recently explored \cite{9711249,294490,tao2025gcpguardedcollaborativeperception, Li_2023_ICCV, zhao2024maliciousagentdetectionrobust}, highlighting security challenges.

Existing works have explored various approaches to exploit these vulnerabilities in CP systems. For feature-level CP, Tu \textit{et al.} \cite{9711249} pioneered feature map perturbations but created obvious modifications easily detectable by anomaly detection. Tao \textit{et al.} \cite{tao2025gcpguardedcollaborativeperception} introduced blind area confusion attacks using victim view information, but remained untargeted and vulnerable to collective validation. Zhang \textit{et al.} \cite{294490} developed targeted attacks with specialized loss functions and LiDAR ray-casting, but did not leverage the systematic optimization of attack timing and location for more effective attacks.

To counter such threats, as shown in Fig. \ref{fig:scenario}, current defensive CP systems typically incorporate validation stages before fusion and perception decision making. Early defenses like ROBOSAC \citep{Li_2023_ICCV}, CP-Guard \citep{hu2024cpguardmaliciousagentdetection}, and MADE \citep{zhao2024maliciousagentdetectionrobust} focus on output-level consensus by cross-validating detection results among CAVs. While effective against simpler attacks, these methods struggle with subtle adversarial perturbations. More advanced defenses like CAD \citep{294490} enhance validation by exchanging supplementary data such as occupancy maps for grid-based verification. 

However, current defensive CP systems remain vulnerable to more sophisticated attacks that could exploit two critical weaknesses. \ding{182} \textit{First, existing defenses lack robustness against attacks with systematic timing and region optimization.} Current attacks do not systematically determine when and where to launch fabrication attacks, while attackers could potentially develop more targeted strategies against consensus-based validation among multiple CAVs. \ding{183} \textit{Second, CP defenses inadvertently disclose vulnerability knowledge that attackers could exploit.} 
Shared collaboration data (e.g., feature maps) and validation data (e.g., occupancy maps) exchanged among CAVs naturally embed implicit confidence information about surrounding environments. Sophisticated attackers could leverage such information to identify regions of collective uncertainty and optimize attack strategies based on evolving confidence patterns. This exploitation enables more adaptive and generalizable attacks across different defensive configurations. 

In this paper, we introduce \texttt{MVIG}, a novel adaptive adversarial CP framework that demonstrates how adversaries could exploit these weaknesses by modeling the vulnerability knowledge disclosed by different defensive CP systems into a \textbf{unified mutual view information graph (MVIG) representation}. By learning the temporal evolution of this graph, adversaries can strategically identify optimal attack locations and timing opportunities, enabling attacks with enhanced adaptiveness and stronger generalizability across various defensive configurations. 

The main contributions of this work are three-fold:
\begin{itemize}
 \item We propose a unified mutual view information graph representation that captures vulnerability disclosed by different CP defense systems, enabling systematic analysis of collective uncertainty patterns and optimal attack opportunities across various defensive configurations.
 \item We propose a novel \texttt{MVIG} attack framework that leverages temporal MVIG learning to generate dynamic fabrication risk maps for identifying vulnerable view regions, along with an entropy-aware vulnerability search to optimize attack timing, location, and persistence.
 \item We conduct comprehensive evaluations showing that \texttt{MVIG} attack significantly outperforms existing methods by decreasing defensive success rates by up to 62\% and maintaining 47\% improved defense evasion for multi-frame attacks, while achieving real-time performance of 29.9 FPS in CP systems.
\end{itemize}

%% file: sections/related_works.tex
\section{Related Works}
\label{sec:related_works}

\noindent \textbf{Robust Collaborative Perception.} Previous works have explored robust CP systems that handle challenges including synchronization issues \citep{lei2022latencyawarecollaborativeperception, NEURIPS2023_5a829e29}, communication interruption \citep{ren2024interruptionawarecooperativeperceptionv2x}, data corruption \citep{zhang2025dsrclearningdensityinsensitivesemanticaware}, and sensor failures \citep{NEURIPS2024_27e5626c}. While these system-level issues can degrade performance, they are often predictable and addressable through proper system design. In contrast, malicious agent threats represent fundamentally different challenges: attackers can deliberately exploit system vulnerabilities through targeted information manipulation, requiring defense mechanisms beyond traditional robustness approaches.

\noindent \textbf{Adversarial Attacks on CP.} Security threats have evolved from physical attacks \citep{9710897, 9519442} to sophisticated feature-level perturbations. Tu \textit{et al.} \citep{9711249} pioneered feature map perturbations with high success rates but obvious modifications easily detectable by anomaly detection \citep{Li_2023_ICCV,zhao2024maliciousagentdetectionrobust}. Tao \textit{et al.} \citep{tao2025gcpguardedcollaborativeperception} improved stealthiness through blind area confusion (BAC) attacks using victim view information, but relied solely on single-agent knowledge without considering collective validation. Zhang \textit{et al.} \citep{294490} advanced targeted attacks with LiDAR ray-casting for real-time perturbation, but lacked systematic optimization of attack timing and location. These limitations motivate our \texttt{MVIG} framework, which exploits mutual view relationships for maximizing attack success.

\noindent \textbf{Defensive CP Systems.} Defense mechanisms have evolved from output-level validation to spatial verification. Early defenses, including ROBOSAC \citep{Li_2023_ICCV}, CP-Guard \citep{hu2024cpguardmaliciousagentdetection}, and MADE \citep{zhao2024maliciousagentdetectionrobust}, verify perception outputs through consensus checking using Hungarian matching and reconstruction loss. While effective against naive attacks, they struggle with sophisticated perturbations. More advanced defenses like CAD \citep{294490} exchange occupancy maps for grid-based verification: flagging inconsistencies in ego-known regions and relying on consensus in unknown regions. However, these defenses inadvertently expose collective perception coverage, enabling attackers to optimize their strategies. Our work reveals fundamental vulnerabilities across both paradigms through systematic analysis of mutual view information gaps.

%% file: sections/method.tex
\begin{figure*}[t]
    \vspace{-5mm}
    \centering
    \includegraphics[width=0.95\linewidth]{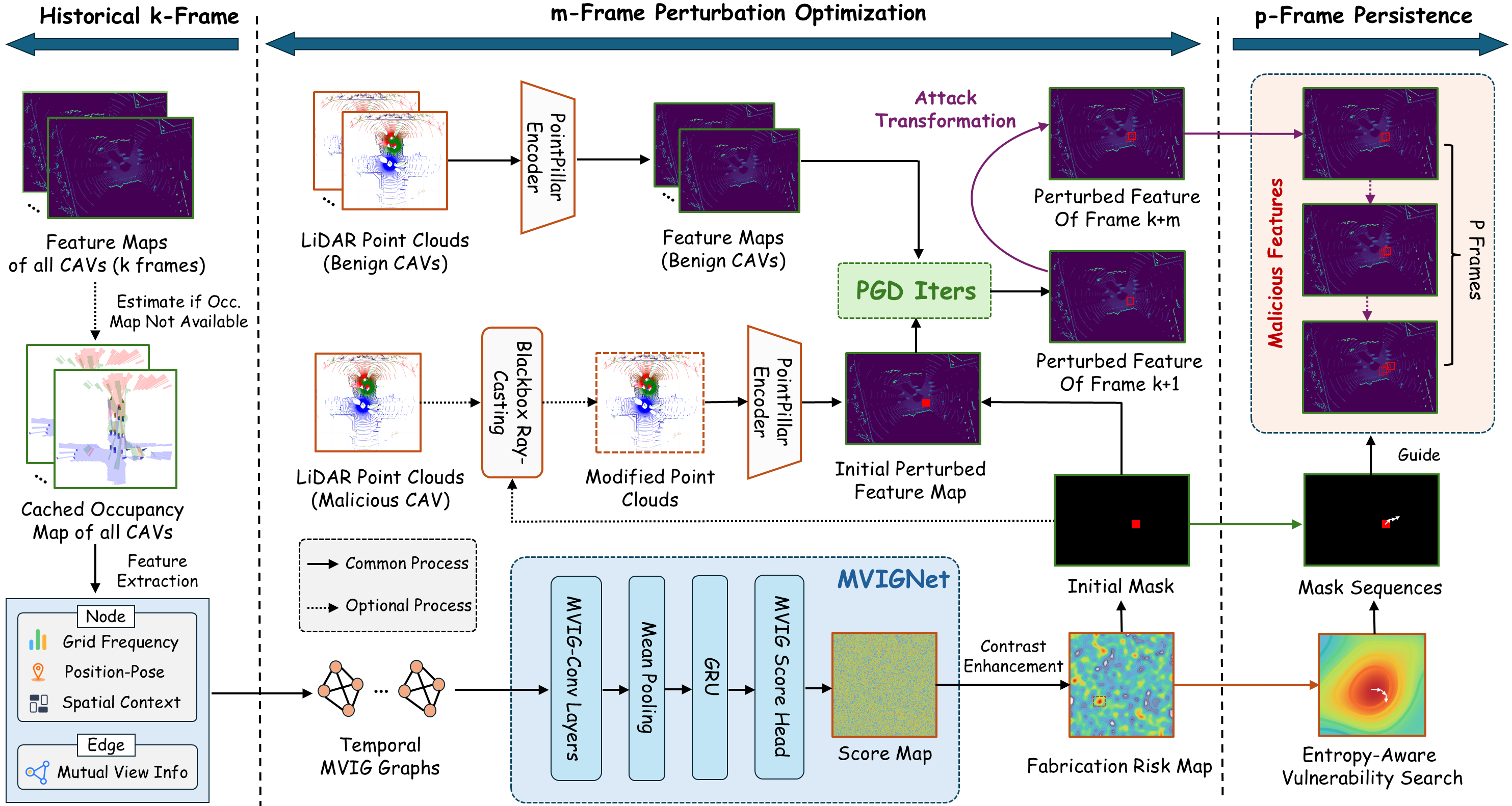}
    \caption{Overview of our \texttt{MVIG} attack framework. It begins with transforming historical CAV collaboration data into mutual view information graphs, generates fabrication risk maps via MVIGNet and employs entropy-aware vulnerability search to optimize both attack timing and location, creating persistent, believable fabricated objects in the victim's perception across consecutive frames.
    }
    \label{fig:mvig_attack}
    \vspace{-5mm}
\end{figure*}

\section{Attack Methodology}

\subsection{Attack Formulation}
We consider a system with $N$ CAVs (denoted as $\mathcal{N}$) jointly executing collaborative perception. At time $t$, the $i$-th CAV captures observation $\mathbf{O}_i$ and generates feature map $\mathbf{F}_i = \Phi_\mathtt{enc}(\mathbf{O}_i)$, where $\Phi_\mathtt{enc}$ is the encoder. After receiving feature maps $\{\mathbf{F}_m\}_{m=1}^{N-1}$ from all CAVs, the normal perception output is $\mathbf{Y}_i = \Phi_\mathtt{dec}\left(f_\mathtt{agg}\left(\{\mathbf{F}_m^t\}_{m=0}^{N-1}\right)\right)$,
where $\Phi_\mathtt{dec}$ is the decoder and $f_\mathtt{agg}$ is the aggregation function.
In our attack scenario, a malicious agent $a$ manipulates perception by transmitting $\mathbf{F}_a^t + \boldsymbol{\delta}$ instead of $\mathbf{F}_a^t$, resulting in a modified perception output $\mathbf{Y}'_i = \Phi_\mathtt{dec}\left(f_\mathtt{agg}\left(\{\mathbf{F}_m^t\}_{m\neq a}, \mathbf{F}_a^t + \boldsymbol{\delta}\right)\right)$.
The attacker caches $k$ historical feature maps from all CAVs, and determines perturbation $\boldsymbol{\delta}$ through:
\begin{equation}
    \boldsymbol{\delta} = \Psi(\{\mathbf{M}_m\}_{m \in \mathcal{N}}), \quad \mathbf{M}_m = \{\mathbf{F}_m^t, \mathbf{F}_m^{t-1}, \ldots, \mathbf{F}_m^{t-k}\},
\end{equation}
where $\Psi(\cdot)$ transforms historical messages into optimal perturbations.
Given attack objective $\mathcal{L}$ and constraints $\mathcal{C}$, the attacker solves:
\begin{equation}
    \max_{\boldsymbol{\delta}} \mathcal{L}(\mathbf{Y}'_i, \mathbf{Y}_i) \quad \text{s.t.} \quad \mathcal{C}(\boldsymbol{\delta}),
    \label{eq:objective}
\end{equation}
This formulation generalizes existing approaches: when $k=0$, it reduces to Tu \textit{et al.}'s model \citep{9711249} using only current frame features; when $k=1$, it matches Zhang \textit{et al.}'s framework \citep{294490} with one historical frame. This unified framework enables systematic comparison of different attack strategies.

\subsection{Adversarial Threat Model}

To construct a fair comparison with existing attack methods, we follow the threat model established by previous CP security works~\cite{294490,Li_2023_ICCV,tao2025gcpguardedcollaborativeperception}, which assume that attackers have compromised at least one legitimate CAV and gained access to the shared CP model parameters. 

The rationale for this assumption lies in the practical deployment of CP systems: models are typically trained centrally by automotive manufacturers or service providers, then distributed to all participating CAVs to ensure consistent feature extraction and fusion. All CAVs in the network use shared model architectures and parameters for proper feature alignment and collaborative decoding. Once a vehicle is compromised, whether through software vulnerabilities, supply chain attacks, or malicious insiders, the attacker naturally inherits the vehicle's legitimate access rights to these shared resources.

Under this standard threat model, we consider an internal attacker that has access to: (1) the shared CP model architecture and parameters, (2) feature maps transmitted among vehicles during collaboration, and (3) occupancy maps if exchanged for defense validation. This access is inherent to the compromised vehicle's legitimate participation in the CP system before being identified and isolated by defense mechanisms.


\subsection{MVIG-Based Adaptive Fabrication Attack}

In this subsection, we introduce the details of our MVIG-based adaptive fabrication attack.

\noindent \textbf{Mutual View Information Graph Construction.}
As illustrated in Fig. \ref{fig:mvig_attack}, our MVIG construction depends on occupancy information from CAVs. Different defensive CP systems disclose varying degrees of this information, and our method adapts accordingly. In systems with CAD defenses, occupancy grid maps are shared among CAVs, allowing attackers direct access to such information. For systems without map sharing, we use the Blind Region Segmentation (BRS) algorithm \citep{tao2025gcpguardedcollaborativeperception} to generate approximate grids from exchanged perception data (Appendix \ref{app:bsc_attack}). This approximation suffices since CP systems without explicit occupancy map sharing inherently have limited defenses against targeted fabrication attacks.

Let $G = (V, E, \mathbf{W}) \in \mathcal{G}$ denote the weighted undirected graph, where $\mathcal{G}$ represents the space of all possible MVIGs. For each node $v_i \in V$ corresponding to CAV $i$, we define the occupancy matrix $O_i \in \{0,1,2\}^{H \times W}$ with states free, occupied, and unknown, where $H,W \in \mathbb{N}$ denote the height and width of the grid, respectively.
For each node $v_i$, we construct a feature vector comprising three complementary components that capture different aspects of the CAV's perception. The basic occupancy distribution component provides a global view of environmental understanding through normalized frequencies of free, occupied, and unknown states across the grid as $\mathbf{h}_i^{\text{basic}} \in \mathbb{R}^{d_1}$, where $\mathbf{h}_i^{\text{basic}} = [p_i(0), p_i(1), p_i(2)]^T$, and the probability of each occupancy state is calculated as:
\begin{equation}
    p_i(c) = \frac{1}{HW}\sum_{x,y} \mathbb{I}[O_i(x,y)=c], c \in \{0,1,2\},
\end{equation}
where $\mathbb{I}[\cdot]$ is the indicator function, and $(x,y)$ are the grid coordinates.
The position-pose information $\mathbf{h}_i^{\text{pos}} \in \mathbb{R}^{d_2}$ encodes the CAV's location and orientation. 
Finally, spatial context features provide rich spatial patterns that capture local and global perception structures as $\mathbf{h}_i^{\text{spatial}} = \Phi_{\text{pool}}(O_i) \in \mathbb{R}^{d_3}$, where $\Phi_{\text{pool}}$ represents multi-scale pooling operations. These features help identify regions with potential detection gaps or overlaps. The complete node feature vector is expressed as $\mathbf{h}_i = [\mathbf{h}_i^{\text{basic}}; \mathbf{h}_i^{\text{pos}}; \mathbf{h}_i^{\text{spatial}}] \in \mathbb{R}^{d_1 + d_2 + d_3} $. 
The edge weight quantifies position-wise perception agreement using mutual information, averaging the overlap across all grid positions:
\begin{equation}\label{eq:edge_weight}
\mathbf{W}_{ij} = \mathbb{E}_{(x,y)}\left[\sum_{a,b=0}^{2} p_{ij}(a,b) \log\frac{p_{ij}(a,b)}{p_i(a)p_j(b)}\right],
\end{equation}
where $\mathcal{I}(\cdot;\cdot)$ denotes the mutual information operator.
Here, $p_{ij}(a,b)$ represents the joint probability of occupancy states at position $(x,y)$, and $p_i(a)$, $p_j(b)$ are the corresponding marginal probabilities. High mutual information strengthens message passing between CAVs with consistent perceptions, while low values reduce influence between CAVs with conflicting observations. This differential flow regulation enables the unified MVIG representation to effectively reveal vulnerabilities exposed by different CP defense systems through both system-level spectral signatures (information flow capacity and consensus fragility) and region-level entropy deficits\footnote{Detailed theoretical analysis of why MVIG can effectively capture these vulnerabilities is provided in Appendix \ref{app:analysis_mvig_representation}.}.

\noindent \textbf{Mask Prediction via MVIGNet.}
Given a sequence of $k$-frame historical MVIGs $\mathcal{G}^{t-k+1:t} = \{G_{t-k+1}, \ldots, G_t\}$, MVIGNet predicts vulnerable regions for future frame $t+m$ through three modules: spatial encoding, temporal modeling, and score generation (detailed in Appendix \ref{app:mvig_structure}).
For spatial encoding, we apply graph convolution layers aggregating neighbor features weighted by mutual view information:
\begin{equation}\label{eq:gnn}
    \mathbf{h}_i^{(l,\tau)} = \sigma\left(\mathbf{\Theta}^{(l)}\mathbf{h}_i^{(l-1,\tau)} + \sum_{j \in \mathcal{N}(i)} \mathbf{h}_j^{(l-1,\tau)} \odot \mathbf{\Psi}_{e}(\mathbf{W}_{ij}^{\tau})\right),
\end{equation}
where $\mathbf{h}_i^{(l,\tau)}$ represents node $i$'s features at layer $l$ and time $\tau$, and $\mathbf{\Psi}_{e}(\cdot)$ transforms edge weights to capture CAV relationship importance. After mean pooling across nodes, we obtain frame-level representations $\smash{\mathbf{f}^{\tau} = \frac{1}{|V|}\sum_{i \in V} \mathbf{h}_i^{(L,\tau)}}$.
For temporal modeling, a GRU extracts patterns from the historical sequence to predict future state:
\begin{equation}
    \mathbf{z}^{t+m} = \boldsymbol{\Phi}_{\text{GRU}}\left(\left\{\mathbf{f}^{\tau}\right\}_{\tau=t-k+1}^{t}\right).
\end{equation}
Finally, a two-layer FCN score head with ReLU activation generates the vulnerability score map:
\begin{equation}\label{eq:score_map}
    \mathbf{S}_{t+m} = \operatorname{Softmax}(\text{FCN}(\mathbf{z}^{t+m})) \in [0,1]^{H \times W}.
\end{equation}
We apply physical constraints (map topology and proximity to victim) to ensure plausible attack locations, then select optimal positions via probability sampling. During inference, test-time augmentation with Gaussian kernel enhancement generates the final fabrication risk map $\tilde{\mathbf{S}}_{t+m}$ (Detailed in Appendix \ref{app:attack_risk_map}).

\definecolor{goodcolor}{rgb}{0.0,0.6,0.0}           
\definecolor{badcolor}{rgb}{0.8,0.0,0.0}            
\begin{table*}[t]
  \caption{Performance comparison of different CP attacks on Adv-OPV2V dataset under various defenses. \textbf{Bold}: best (lowest DSR), \underline{Underline}: second-best. \textcolor{goodcolor}{Green}: improvement over second-best, \textcolor{badcolor}{Red}: degradation. Spoof methods compete within spoof group, remove methods within remove group.}
  \centering
  \setlength{\tabcolsep}{5pt}
  \renewcommand{\arraystretch}{1.5}
  \definecolor{lightgray}{rgb}{0.965,0.965,0.965}     
  \definecolor{highlightcolor}{rgb}{0.965,0.98,1.0}   
  \resizebox{\textwidth}{!}{%
  \footnotesize
  \begin{tabular}{l|cc|ccc|ccc|ccc|ccc}
  \hline
  \multirow{2}{*}{\textbf{Methods}} & \multicolumn{2}{c|}{\textbf{No Defense}} & \multicolumn{3}{c|}{\textbf{ROBOSAC \citep{Li_2023_ICCV}}} & \multicolumn{3}{c|}{\textbf{CP-Guard \citep{hu2024cpguardmaliciousagentdetection}}} & \multicolumn{3}{c|}{\textbf{GCP \citep{tao2025gcpguardedcollaborativeperception}}} & \multicolumn{3}{c}{\textbf{CAD \citep{294490}}} \\
  \cline{2-15}
  & \textbf{ASR} & \textbf{$\Delta$AP} & \textbf{DSR} & \textbf{TPR} & \textbf{FPR} & \textbf{DSR} & \textbf{TPR} & \textbf{FPR} & \textbf{DSR} & \textbf{TPR} & \textbf{FPR} & \textbf{DSR} & \textbf{TPR} & \textbf{FPR} \\
  \hline
  Basic \cite{9711249} & 100.0 & -62.7 & 100.0 & 100.0 & 1.9 & 100.0 & 100.0 & 2.8 & 100.0 & 100.0 & 1.8 & 100.0 & 71.3 & 2.2 \\
  \hline
  RC (spoof) \cite{294490} & 92.4 & -1.9 & \textbf{12.0} & 6.3 & 1.4 & \underline{18.1} & 9.6 & 2.4 & \textbf{12.2} & 6.4 & 1.8 & \underline{83.5} & 71.2 & 0.8 \\
  RC  (remove) \cite{294490} & 98.0 & -3.6 & \underline{14.2} & 7.2 & 1.6 & \underline{21.4} & 11.0 & 1.7 & \underline{15.0} & 8.1 & 1.9 & \underline{90.1} & 72.0 & 1.0 \\
  \hline
  BAC  \cite{tao2025gcpguardedcollaborativeperception} & 99.2 & -16.7 & 23.0 & 13.7 & 1.8 & 37.0 & 22.7 & 2.8 & 28.0 & 16.6 & 1.8 & 90.5 & 78.1 & 1.8 \\
  \hline
  \textbf{\texttt{MVIG} (spoof)} & 94.3 & -1.8 & \multicolumn{1}{l}{\underline{14.8}\textsuperscript{\textcolor{badcolor}{+23\%}}} & 7.9 & 1.3 & \multicolumn{1}{l}{\textbf{17.2}\textsuperscript{\textcolor{goodcolor}{-5\%}}} & 9.3 & 2.5 & \multicolumn{1}{l}{\underline{13.0}\textsuperscript{\textcolor{badcolor}{+7\%}}} & 6.9 & 0.9 & \multicolumn{1}{l}{\textbf{32.0}\textsuperscript{\textcolor{goodcolor}{-62\%}}} & 58.7 & 9.1 \\
  \textbf{\texttt{MVIG} (remove)} & 96.8 & -4.2 & \multicolumn{1}{l}{\textbf{12.2}\textsuperscript{\textcolor{goodcolor}{-14\%}}} & 6.3 & 1.7 & \multicolumn{1}{l}{\textbf{21.3}\textsuperscript{\textcolor{goodcolor}{-0\%}}} & 11.5 & 1.2 & \multicolumn{1}{l}{\textbf{14.2}\textsuperscript{\textcolor{goodcolor}{-5\%}}} & 7.5 & 1.9 & \multicolumn{1}{l}{\textbf{78.2}\textsuperscript{\textcolor{goodcolor}{-13\%}}} & 73.6 & 1.4 \\
  \hline
  \end{tabular}%
  }
  \label{tab:attack_performance}
\end{table*}

\noindent \textbf{Alternating Optimization Strategy.}
Our attack employs a decoupled two-stage optimization: given mask $M_t^*$ from MVIGNet, we optimize feature perturbation $\boldsymbol{\delta}$ via PGD \citep{madry2018towards}; then MVIGNet learns optimal mask positions using the perturbed results.
For feature perturbation, PGD maximizes the objective:
\begin{equation}
\phi(\boldsymbol{\delta}, M_t^*) = 
\begin{cases}
\sum_{b \in B'} \text{IoU}(b, b_t) \cdot \log(b_{\sigma}), & \text{spoofing} \\
-\sum_{b \in B'} \text{IoU}(b, b_t) \cdot \log(b_{\sigma}), & \text{removal}
\end{cases}
\end{equation}
where $B'$ denotes detected boxes after perturbation, $b_{\sigma}$ is box confidence, and $b_t$ is the target box. Mask $M_t^*$ constrains perturbations to vulnerable regions, maintaining stealthiness.

For mask optimization, MVIGNet trains with a multi-objective loss (detailed in Appendix \ref{app:loss_functions}): \textit{(1) Attack effectiveness loss} maximizes fabrication confidence while penalizing distance from victim; \textit{(2) Box differentiation loss} minimizes overlap with existing detections; \textit{(3) Defense evasion loss} avoids conflict regions identified from multi-CAV occupancy maps where contradictory observations could trigger defenses.

Crucially, the two stages are decoupled: MVIGNet uses PGD-generated boxes to compute loss without gradient backpropagation to PGD. At inference, MVIGNet directly predicts masks while PGD iteratively refines perturbations.

\noindent \textbf{Online Attack and Attack Persistence.}
To enable online attack, we implement an $m$-frame delay mechanism where MVIGNet predicts optimal attack positions $m$ frames ahead, balancing prediction accuracy with operational timing. Based on overhead analysis in Appendix \ref{app:overhead_analysis}, we set $m=2$. At execution, we directly apply precomputed perturbations through feature mapping between feature map coordinates and 3D real-world space, exploiting feature stability across consecutive frames.
For attack persistence, we extend a single optimization across $p$ consecutive frames by pre-computing masks based on initial fabrication risk map:
\begin{equation}\label{eq:mask_transform}
    \mathbf{M}_{t+m+j} = \mathcal{T}_j(\mathbf{M}_{t+m}, \tilde{\mathbf{S}}_{t+m}), \quad j \in \{1,2,\ldots,p-1\}.
\end{equation}
The transformation operator $\mathcal{T}_j$ employs entropy-aware vulnerability search (Appendix \ref{app:entropy_aware_search}) to ensure temporal consistency, while attack continuation is determined by:
\begin{equation}\label{eq:continue_attack}
    \mathcal{C}_{t+m+j} = \mathbb{I}\left[\mathbb{E}_{(x,y) \in \mathcal{N}(x_c,y_c)}\left[\tilde{\mathbf{S}}_{t+m}(x,y)\right] \geq \eta\right],
\end{equation}
where $\mathcal{C}_{i}$ indicates attack continuation in frame $i$, $\mathbb{E}$ denotes the expectation operator, $\mathcal{N}(x_c,y_c)$ is the neighborhood around mask center $(x_c,y_c)$, and $\eta$ is the continuation threshold.
For frames within the persistence window, we warp previously generated perturbations to new positions, with limited duration to avoid detection risk from outdated information.

%% file: sections/experiments.tex

\section{Experiments}

\definecolor{goodcolor}{rgb}{0.0,0.6,0.0}           
\definecolor{badcolor}{rgb}{0.8,0.0,0.0}            
\begin{table*}[t]
  \caption{Performance comparison of different CP attacks with various persistence. \textbf{Bold}: best (lowest DSR), \underline{Underline}: second-best. \textcolor{goodcolor}{Green}: improvement over second-best, \textcolor{badcolor}{Red}: degradation. Spoof attack compete within spoof group, remove attack within remove group.}
  \centering
  \setlength{\tabcolsep}{5pt}
  \renewcommand{\arraystretch}{1.1}
  \definecolor{lightgray}{rgb}{0.965,0.965,0.965}     
  \definecolor{highlightcolor}{rgb}{0.965,0.98,1.0}   
  \resizebox{\textwidth}{!}{%
  \footnotesize
  \begin{tabular}{l|c|ccc|ccc|ccc|ccc}
  \hline
  \multirow{2}{*}{\textbf{Methods}} & \multirow{2}{*}{\textbf{Persistence}} & \multicolumn{3}{c|}{\textbf{ROBOSAC+ \citep{Li_2023_ICCV}}} & \multicolumn{3}{c|}{\textbf{CP-Guard \citep{hu2024cpguardmaliciousagentdetection}}} & \multicolumn{3}{c|}{\textbf{GCP \citep{tao2025gcpguardedcollaborativeperception}}} & \multicolumn{3}{c}{\textbf{CAD \citep{294490}}} \\
  \cline{3-14}
  & & \textbf{DSR} & \textbf{TPR} & \textbf{FPR} & \textbf{DSR} & \textbf{TPR} & \textbf{FPR} & \textbf{DSR} & \textbf{TPR} & \textbf{FPR} & \textbf{DSR} & \textbf{TPR} & \textbf{FPR} \\
  \hline
  \multirow{3}{*}{Basic \cite{9711249}} 
  & 0-frame & 100.0 & 100.0 & 1.9 & 100.0 & 100.0 & 2.8 & 100.0 & 100.0 & 1.8 & 100.0 & 71.3 & 2.2 \\
  & 1-frame & 100.0 & 100.0 & 3.7 & 100.0 & 100.0 & 2.8 & 100.0 & 100.0 & 5.8 & 100.0 & 75.1 & 2.0 \\
  & 3-frame & 100.0 & 100.0 & 2.8 & 100.0 & 100.0 & 0.9 & 100.0 & 100.0 & 0.6 & 100.0 & 76.2 & 2.0 \\
  \hline
  \multirow{3}{*}{RC (spoof) \cite{294490}} 
  & 0-frame & \textbf{12.0} & 6.3 & 1.4 & \underline{18.1} & 9.6 & 2.4 & \textbf{12.2} & 6.4 & 1.8 & \underline{83.5} & 71.2 & 0.8 \\
  & 1-frame & \underline{29.1} & 8.2 & 3.7 & \textbf{19.2} & 5.1 & 1.2 & \underline{37.2} & 11.3 & 5.8 & \underline{98.2} & 76.9 & 1.5 \\
  & 3-frame & \underline{38.1} & 7.7 & 2.8 & \textbf{24.0} & 4.5 & 1.1 & \underline{42.1} & 9.1 & 5.9 & \underline{100.0} & 74.4 & 1.2 \\
  \hline
  \multirow{3}{*}{RC (remove) \cite{294490}} 
  & 0-frame & \underline{14.2} & 7.2 & 1.6 & \underline{21.4} & 11.0 & 1.7 & \underline{15.0} & 8.1 & 1.9 & \underline{90.1} & 72.0 & 1.0 \\
  & 1-frame & \textbf{28.1} & 8.0 & 4.1 & \textbf{24.1} & 7.7 & 1.2 & \textbf{40.0} & 12.2 & 5.3 & \underline{91.1} & 73.9 & 0.2 \\
  & 3-frame & \underline{39.2} & 8.3 & 3.4 & \underline{34.2} & 7.1 & 1.1 & \underline{48.2} & 11.2 & 6.4 & \underline{98.0} & 72.2 & 1.2 \\
  \hline
  \multirow{3}{*}{BAC \cite{tao2025gcpguardedcollaborativeperception}} 
  & 0-frame & 23.0 & 13.7 & 1.8 & 37.0 & 22.7 & 2.8 & 28.0 & 16.6 & 1.8 & 90.5 & 78.1 & 1.8 \\
  & 1-frame & 58.1 & 35.5 & 1.7 & 53.2 & 18.9 & 1.2 & 76.2 & 46.7 & 2.0 & 92.2 & 83.0 & 2.0 \\
  & 3-frame & 74.2 & 40.5 & 1.5 & 68.1 & 20.8 & 1.1 & 86.0 & 57.5 & 2.2 & 98.0 & 82.1 & 2.1 \\
  \hline
  \multirow{3}{*}{\textbf{\texttt{MVIG} (spoof)}} 
  & 0-frame & \multicolumn{1}{l}{\underline{14.8}\textsuperscript{\textcolor{badcolor}{+23\%}}} & 7.9 & 1.3 & \multicolumn{1}{l}{\textbf{17.2}\textsuperscript{\textcolor{goodcolor}{-5\%}}} & 9.3 & 2.5 & \multicolumn{1}{l}{\underline{13.0}\textsuperscript{\textcolor{badcolor}{+7\%}}} & 6.9 & 0.9 & \multicolumn{1}{l}{\textbf{32.0}\textsuperscript{\textcolor{goodcolor}{-62\%}}} & 58.7 & 9.1 \\
  & 1-frame & \multicolumn{1}{l}{\textbf{29.0}\textsuperscript{\textcolor{goodcolor}{-0\%}}} & 8.2 & 3.7 & \multicolumn{1}{l}{\underline{22.0}\textsuperscript{\textcolor{badcolor}{+15\%}}} & 6.1 & 1.2 & \multicolumn{1}{l}{\textbf{37.0}\textsuperscript{\textcolor{goodcolor}{-1\%}}} & 11.0 & 5.8 & \multicolumn{1}{l}{\textbf{48.2}\textsuperscript{\textcolor{goodcolor}{-51\%}}} & 60.4 & 1.5 \\
  & 3-frame & \multicolumn{1}{l}{\textbf{37.0}\textsuperscript{\textcolor{goodcolor}{-3\%}}} & 7.4 & 2.8 & \multicolumn{1}{l}{\underline{26.2}\textsuperscript{\textcolor{badcolor}{+9\%}}} & 4.9 & 1.1 & \multicolumn{1}{l}{\textbf{42.0}\textsuperscript{\textcolor{goodcolor}{-0\%}}} & 9.1 & 6.1 & \multicolumn{1}{l}{\textbf{63.2}\textsuperscript{\textcolor{goodcolor}{-37\%}}} & 52.5 & 1.2 \\
  \hline
  \multirow{3}{*}{\textbf{\texttt{MVIG} (remove)}} 
  & 0-frame & \multicolumn{1}{l}{\textbf{12.2}\textsuperscript{\textcolor{goodcolor}{-14\%}}} & 6.3 & 1.7 & \multicolumn{1}{l}{\textbf{21.3}\textsuperscript{\textcolor{goodcolor}{-0\%}}} & 11.5 & 1.2 & \multicolumn{1}{l}{\textbf{14.2}\textsuperscript{\textcolor{goodcolor}{-5\%}}} & 7.5 & 1.9 & \multicolumn{1}{l}{\textbf{78.2}\textsuperscript{\textcolor{goodcolor}{-13\%}}} & 73.6 & 1.4 \\
  & 1-frame & \multicolumn{1}{l}{\underline{32.5}\textsuperscript{\textcolor{badcolor}{+16\%}}} & 9.2 & 4.1 & \multicolumn{1}{l}{\underline{28.0}\textsuperscript{\textcolor{badcolor}{+16\%}}} & 8.1 & 1.6 & \multicolumn{1}{l}{\underline{44.1}\textsuperscript{\textcolor{badcolor}{+10\%}}} & 13.8 & 5.3 & \multicolumn{1}{l}{\textbf{80.4}\textsuperscript{\textcolor{goodcolor}{-12\%}}} & 74.5 & 0.5 \\
  & 3-frame & \multicolumn{1}{l}{\textbf{37.0}\textsuperscript{\textcolor{goodcolor}{-6\%}}} & 7.7 & 3.4 & \multicolumn{1}{l}{\textbf{34.0}\textsuperscript{\textcolor{goodcolor}{-1\%}}} & 7.0 & 0.9 & \multicolumn{1}{l}{\textbf{46.0}\textsuperscript{\textcolor{goodcolor}{-5\%}}} & 10.5 & 6.4 & \multicolumn{1}{l}{\textbf{84.2}\textsuperscript{\textcolor{goodcolor}{-14\%}}} & 73.1 & 1.9 \\
  \hline
  \end{tabular}%
  }
  \label{tab:temporal_attack_performance}
\end{table*}


\begin{figure*}[t]
  \centering
  \includegraphics[width=\linewidth]{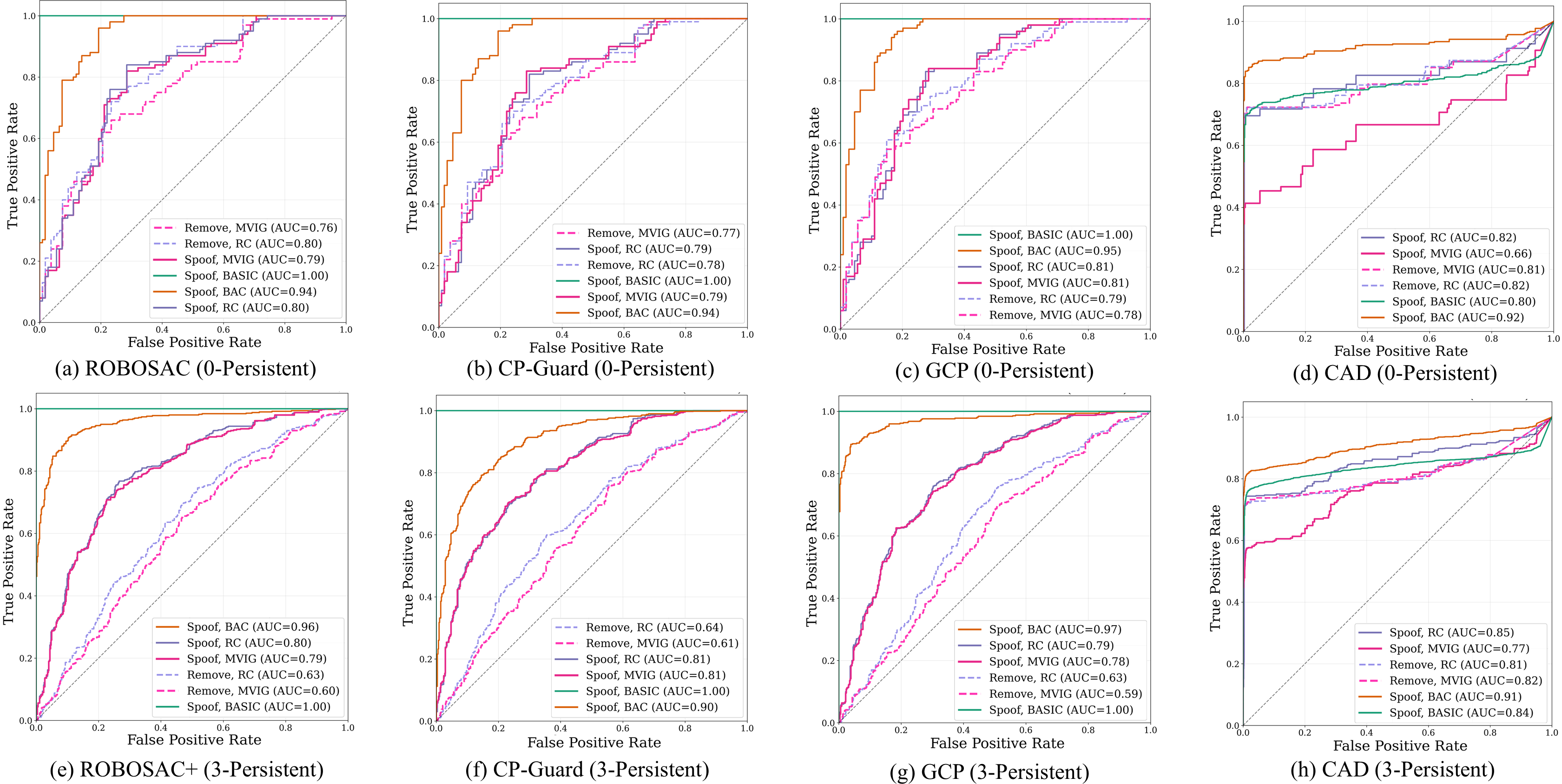}
  \caption{ROC curves of different defenders on various CP attacks with different persistence.}
  \label{fig:roc_defense}
  \vspace{-0.5cm}
\end{figure*}

\begin{figure*}[t]
  \centering
  \includegraphics[width=0.95\linewidth]{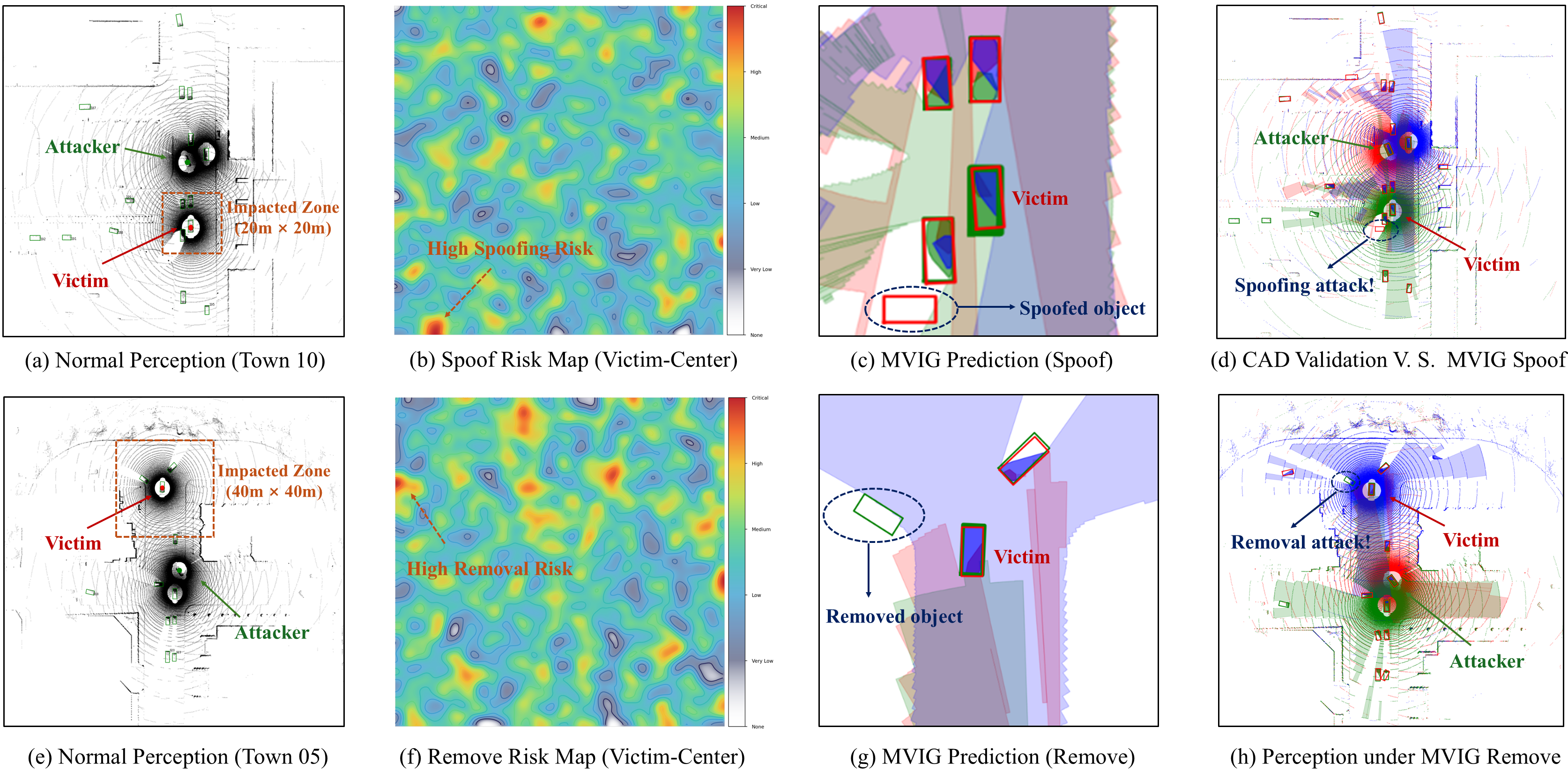}
  \caption{\texttt{MVIG} attack in CP systems. \textcolor{green}{Green} boxes represent predictions without attack, \textcolor{red}{red} boxes show predictions after attack. Color-filled polygon regions in (c,g) correspond to GT occupancy information from different CAVs, dark: occupied, light: free, white: unknown.}
  \label{fig:attack_evolution}
\end{figure*}

\begin{figure*}[t]
  \centering
  \includegraphics[width=0.96\linewidth]{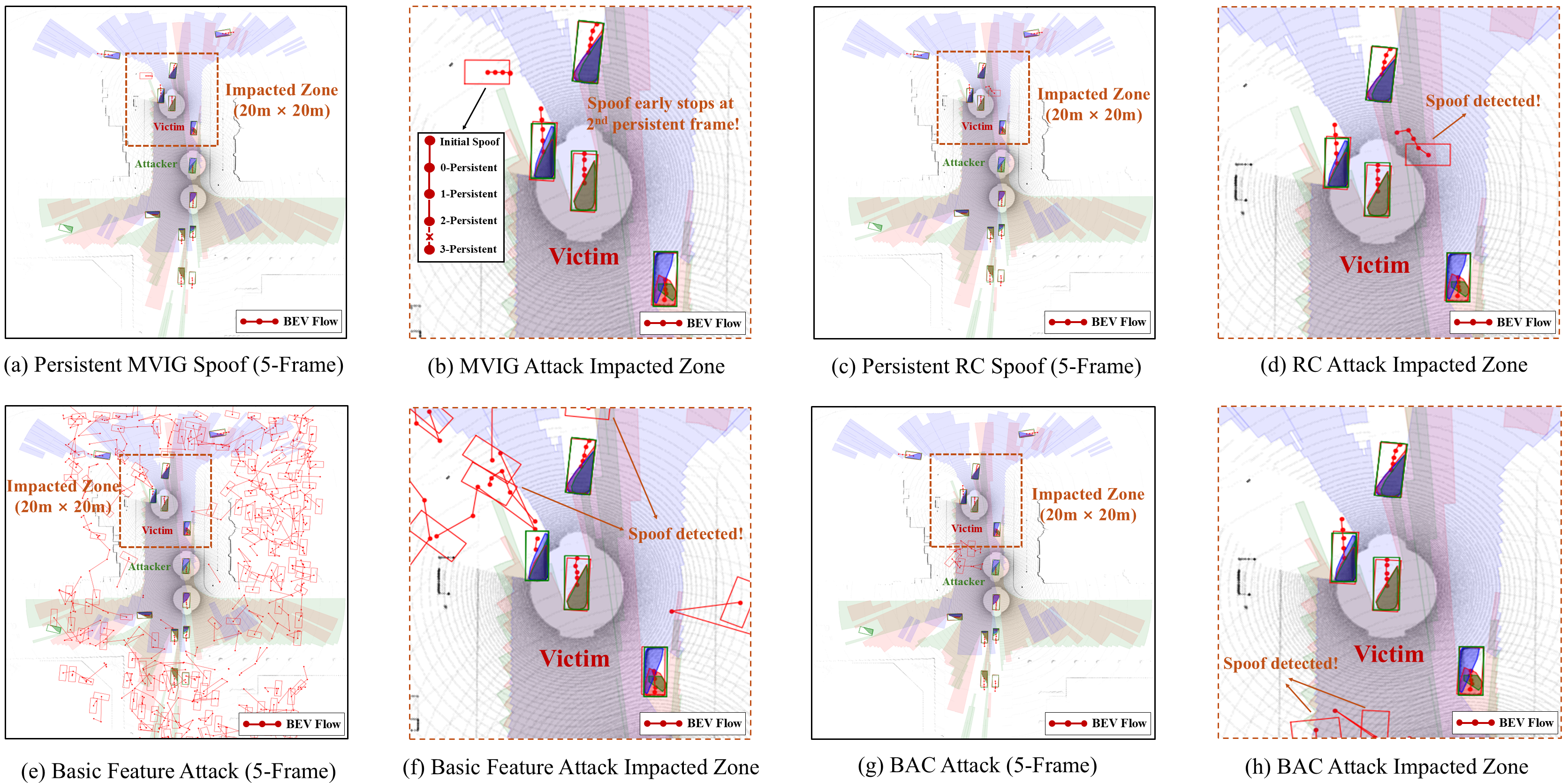}
  \caption{Comparison of persistent attacks and their impacted zones. The BEV flow is calculated by matching the most similar objects in the consecutive detection maps. }
  \label{fig:detection_vis}
  \vspace{-0.5cm}
\end{figure*}

\subsection{Experimental Setup}
\label{sec:experimental_setup}

\textbf{Datasets.}
We use two datasets: (1) OPV2V \cite{xuOPV2VOpenBenchmark2022}, a large-scale multi-vehicle collaborative perception dataset collected via CARLA simulator \citep{Dosovitskiy17}, containing 11,000+ frames from 2-7 connected vehicles with 10Hz sensors (cameras, LiDAR, GPS/IMU). (2) Adv-OPV2V \cite{294490}, the first benchmark for CP attacks and defenses built on OPV2V, with 300 scenarios each for spoofing and removal attacks, comprising 10 consecutive frames with 3-5 CAVs per scenario. We train MVIG on OPV2V scenarios separate from those in Adv-OPV2V and evaluate on the latter.

\noindent \textbf{Attack Settings.}
In training, we randomly select attacker-target vehicle pairs in OPV2V, while using pre-designated pairs by Adv-OPV2V for testing. Attack magnitude is constrained with $\|\delta\|_{\infty} \leq 1.0$, optimized via PGD (maximum number of iterations = 5, step size = 0.01). Targeted attacks restrict perturbations to 5m $\times$ 5m squares, while untargeted attacks have no spatial restrictions.

\noindent \textbf{Implementation Details.}
We train our MVIGNet on 1000 diverse traffic scenarios from OPV2V (80\% training, 20\% validation), using Adam optimizer with learning rate 0.001 for 50 epochs. We set prediction horizon $m=2$ and temporal window size $k=5$. Default size of impacted zone are 20m $\times$ 20m for spoofing and 40m $\times$ 40m for removal attacks, which aligns with previous work \cite{294490} showing that fabricated objects within this range can significantly impact downstream driving decision-making. We use PointPillars \citep{Lang_2019_CVPR} as LiDAR detector and AttnFuse \citep{xuOPV2VOpenBenchmark2022} as CP model.

\noindent \textbf{Baselines and Evaluation Metrics.}
We compare \texttt{MVIG} with three CP attacks (Basic Feature Attack \cite{9711249}, BAC Attack \cite{tao2025gcpguardedcollaborativeperception}, and RC Attack \cite{294490}) and five defensive CP systems (CAD \cite{294490}, ROBOSAC \cite{Li_2023_ICCV}, CP-Guard \cite{hu2024cpguardmaliciousagentdetection}, and GCP \cite{tao2025gcpguardedcollaborativeperception}). We evaluate several metrics: Attack Success Rate (ASR) for undefended scenarios, Defense Success Rate (DSR) for defended scenarios (lower DSR indicates more effective attacks), $\Delta$AP@50 to quantify perception impact, True/False Positive Rates (TPR/FPR) for defender evaluation, and Frames Per Second (FPS) to assess computational overhead. More details are in Appendix~\ref{app:more_implementation_details}.

\subsection{Quantitative Results}

\textbf{Benchmark Comparison.}
Table \ref{tab:attack_performance} reveals that \texttt{MVIG}'s superior performance stems from its ability to translate diverse defenses into unified vulnerability representations. While targeted attacks achieve similar perception impact ($\Delta$AP $\approx$ -2\% to -4\%), \texttt{MVIG} demonstrates substantially lower DSR values, particularly against CAD (0.32 vs. 0.85 for the second-best), because it systematically exploits information asymmetries in mutual view relationships rather than relying on heuristic spatial strategies. This enables \texttt{MVIG} to consistently identify vulnerable regions across different defenses, whether outlier-based (ROBOSAC, CP-Guard, GCP) or occupancy-based (CAD), by capturing fundamental vulnerabilities in CP through spectral graph properties.

\noindent \textbf{Evaluation of Persistent Attack.}
Table \ref{tab:temporal_attack_performance} demonstrates that temporal consistency, not just spatial optimization, is critical for evading multi-frame defenses. The striking contrast between BAC's high detection rates (DSR $>$ 0.98) and \texttt{MVIG}'s maintained effectiveness (DSR = 0.63 at 3-frame persistence) against CAD reveals a fundamental difference: while spatial attacks create inconsistent temporal patterns easily flagged by multi-frame validation, \texttt{MVIG}'s temporal optimization learns to maintain attack coherence across frames by predicting future vulnerability evolution, thereby exploiting temporal blind spots in defense mechanisms.

\noindent \textbf{ROC Analysis.}
Figure \ref{fig:roc_defense} confirms that \texttt{MVIG}'s advantage persists across defense thresholds. The consistently lower AUC values (e.g., 0.77 vs. 0.85-0.91 against CAD with 3-frame persistence) indicate that \texttt{MVIG} fundamentally reduces the separability between malicious and benign transmissions. This robustness stems from attacking regions where multiple CAVs naturally exhibit conflicting observations due to occlusions and view angles, making fabrications appear as legitimate perception differences rather than adversarial anomalies.

\noindent \textbf{Overhead.}
\texttt{MVIG} achieves real-time performance (15.6-29.9 FPS) by decoupling mask prediction from perturbation optimization, ensuring all computations complete before target frames arrive. Details in Appendix \ref{app:overhead_analysis}.

\begin{figure}[t]
  \centering
  \includegraphics[width=1.0\linewidth]{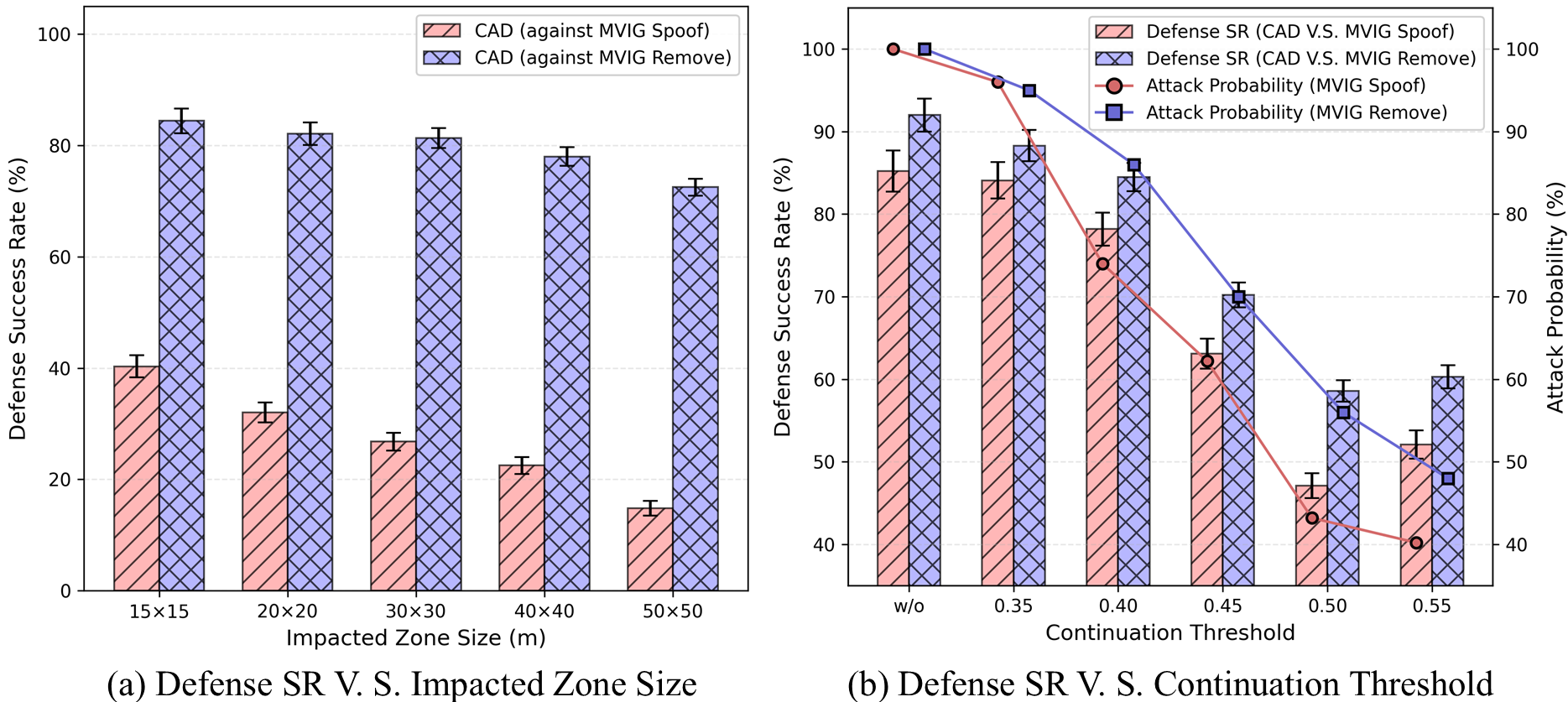}
  \caption{Ablation studies on the impact of impacted zone size and attack continuation threshold.}
  \label{fig:ablation}
  \vspace{-0.3cm}
\end{figure}

\subsection{Qualitative Results}

\noindent \textbf{Visualization of MVIG Attack Effectiveness.}
Figure \ref{fig:attack_evolution} shows our \texttt{MVIG} attack effectiveness in CP systems. These results demonstrate how \texttt{MVIG} identifies optimal attack timing and regions while maintaining plausibility. The predicted optimal attack locations are subsequently reflected in the local occupancy grid maps shown in (c,g), where \texttt{MVIG} attack is prone to exploit the mutually unknown regions to benign CAVs to evade object fabrication detection.

\noindent \textbf{Visualization of Persistent Attack.}
Figure \ref{fig:detection_vis} visualizes different persistent attacks. \texttt{MVIG} attack employs moment-constrained greedy search to optimize attack positions based on fabrication risk maps, achieving superior temporal consistency with smooth BEV flow. \texttt{MVIG} features intelligent attack timing selection with automatic interruption, early stopping at the 2nd persistent frame when the continuation threshold detects low fabrication chances from fabrication risk map.

\subsection{Ablation Studies}

\noindent \textbf{Effect of Impacted Zone Size.}
Figure \ref{fig:ablation}(a) shows how impacted zone size affects attack effectiveness against CAD defense. For \texttt{MVIG} spoof, increasing zone size from 15m to 50m significantly reduces DSR (40.2\% to 14.8\%) by enabling exploration of broader attack regions. \texttt{MVIG} remove shows more stable performance across different zone sizes (DSR between 72.1\% and 84.5\%) due to the constraint of targeting existing objects, limiting the benefit of expanded attack search space.

\noindent \textbf{Effect of Attack Continuation Threshold.}
Figure \ref{fig:ablation}(b) illustrates the impact of attack continuation threshold on defense evasion. Without setting a threshold, attacks face strong defense responses (DSR 85.3\%/91.7\% for \texttt{MVIG} spoof/remove). As threshold increases from 0.35 to 0.55, both attack probability and DSR decrease significantly. This demonstrates that strategic early termination in low fabrication probability regions enhances \texttt{MVIG} attack's defense evasion capabilities, with optimal threshold value between 0.45-0.50 balancing attack opportunity with detection avoidance.

\subsection{Adaptiveness and Generalizability}

Our \texttt{MVIG} attack translates vulnerabilities in CP systems into a unified MVIG representation, which does not rely on specific defense knowledge and can be generalized to any defensive CP system. This enables adaptive attack strategies across two cases:

\noindent \textbf{Case 1: No Defense Knowledge.}
For defensive CP systems without any disclosed defense knowledge (e.g., ROBOSAC, CP-Guard, GCP) or unseen systems, attackers directly estimate MVIG representations from inherently transmitted feature maps to launch attacks. Despite the rough estimation, Table \ref{tab:attack_performance} shows that \texttt{MVIG} attack maintains strong effectiveness with consistently low DSR values (0.12-0.32), demonstrating excellent generalizability.

\noindent \textbf{Case 2: Partial Defense Knowledge.}
When partial defense knowledge is available through exchanged validation data (e.g., occupancy maps in CAD), attackers convert this knowledge into more precise MVIG representations, thereby enhancing attack capabilities. Notably, MVIG achieves 62\% lower DSR against CAD compared to the second-best baseline, demonstrating superior adaptiveness in exploiting available defense information.

This demonstrates the adaptive intelligence of our \texttt{MVIG} attack: it maintains robust effectiveness against systems that lack explicit validation data exchange, while strategically escalating attack strength when such validation data becomes available and exploitable for precise vulnerability localization. In addtion, our \texttt{MVIG} attack is also robust to occupancy estimation errors and different CP architectures. More experimental proofs and discussions on potential countermeasures are provided in Appendix \ref{app:more_experimental_results} and \ref{app:more_discussions}.

%% file: sections/conclusion_ackno.tex
\section{Conclusion}

This paper has introduced \texttt{MVIG} attack, a novel adaptive attack framework that translates vulnerabilities in different defensive CP systems into unified MVIG representations. By systematically exploiting mutual view information asymmetries, \texttt{MVIG} identifies vulnerable regions and optimal attack timing through spectral graph analysis and temporal modeling. Comprehensive evaluations demonstrate \texttt{MVIG} significantly outperforms existing attacks with real-time performance, revealing critical security vulnerabilities in current CP systems.

\noindent \textbf{Limitations and Future Work.}
Our attack has two main limitations: (1) removal attacks have constrained optimization space due to targeting existing objects; (2) effectiveness is bounded by benign CAVs' joint coverage ratio, though strategic timing can exploit dynamic coverage gaps. Future work includes extending the framework to camera-based perception systems by replacing ray-casting with image-based fabrication techniques  and finding effective defense strategies aginst such attacks.


%% file: sections/appendix.tex
\clearpage
\newpage

\appendix

\twocolumn[{
\begin{center}
\Large\textbf{Learning Mutual View Information Graph for Adaptive Adversarial Collaborative Perception}

\vspace{0.3em}

\Large Supplementary Material
\end{center}

\vspace{0.8em}
}]

\noindent\textbf{Contents}

\vspace{0.5em}

\begin{itemize}[leftmargin=0.5cm, label={}]
    \item \textcolor{cvprblue}{\textbf{A~~Analysis of MVIG Representation}} \dotfill {\color{black}\textbf{\pageref{app:analysis_mvig_representation}}}
    \begin{itemize}[leftmargin=0.5cm, label={}]
        \item \textcolor{cvprblue}{A.1~~System-Level Vulnerability Characterization} \dotfill {\color{black}\pageref{subsec:info_theoretic}}
        \item \textcolor{cvprblue}{A.2~~Region-Level Vulnerability Quantification} \dotfill {\color{black}\pageref{subsec:entropy_based}}
    \end{itemize}
    \item \textcolor{cvprblue}{\textbf{B~~Loss Function}} \dotfill {\color{black}\textbf{\pageref{app:loss_functions}}}
    \item \textcolor{cvprblue}{\textbf{C~~More Implementation Details}} \dotfill {\color{black}\textbf{\pageref{app:more_implementation_details}}}
    \begin{itemize}[leftmargin=0.5cm, label={}]
        \item \textcolor{cvprblue}{C.1~~Evaluation Metrics} \dotfill {\color{black}\pageref{app:evaluation_metrics}}
        \item \textcolor{cvprblue}{C.2~~Baselines} \dotfill {\color{black}\pageref{app:baselines}}
    \end{itemize}
    \item \textcolor{cvprblue}{\textbf{D~~More Experimental Results}} \dotfill {\color{black}\textbf{\pageref{app:more_experimental_results}}}
    \begin{itemize}[leftmargin=0.5cm, label={}]
        \item \textcolor{cvprblue}{D.1~~Visualization of Feature Maps} \dotfill {\color{black}\pageref{app:more_experimental_results}}
        \item \textcolor{cvprblue}{D.2~~Results with Different CP Architectures} \dotfill {\color{black}\pageref{app:cp_architectures}}
        \item \textcolor{cvprblue}{D.3~~Scalability Analysis} \dotfill {\color{black}\pageref{app:scalability}}
        \item \textcolor{cvprblue}{D.4~~Robustness to Occupancy Map Errors} \dotfill {\color{black}\pageref{app:robustness}}
    \end{itemize}
    \item \textcolor{cvprblue}{\textbf{E~~More Discussions}} \dotfill {\color{black}\textbf{\pageref{app:more_discussions}}}
    \begin{itemize}[leftmargin=0.5cm, label={}]
        \item \textcolor{cvprblue}{E.1~~Attack Effectiveness Analysis} \dotfill {\color{black}\pageref{app:attack_effectiveness}}
        \item \textcolor{cvprblue}{E.2~~Potential Countermeasures Discussion} \dotfill {\color{black}\pageref{app:countermeasures}}
        \item \textcolor{cvprblue}{E.3~~Communication Delay} \dotfill {\color{black}\pageref{app:communication_delay}}
    \end{itemize}
    \item \textcolor{cvprblue}{\textbf{F~~Overhead Analysis}} \dotfill {\color{black}\textbf{\pageref{app:overhead_analysis}}}
    \begin{itemize}[leftmargin=0.5cm, label={}]
        \item \textcolor{cvprblue}{F.1~~Runtime Performance} \dotfill {\color{black}\pageref{app:overhead_analysis}}
        \item \textcolor{cvprblue}{F.2~~Resource Requirements} \dotfill {\color{black}\pageref{app:overhead_analysis}}
        \item \textcolor{cvprblue}{F.3~~Prediction Horizon Selection} \dotfill {\color{black}\pageref{app:overhead_analysis}}
        \item \textcolor{cvprblue}{F.4~~Component-wise Analysis} \dotfill {\color{black}\pageref{app:overhead_analysis}}
    \end{itemize}
    \item \textcolor{cvprblue}{\textbf{G~~Fabrication Risk Map Generation}} \dotfill {\color{black}\textbf{\pageref{app:attack_risk_map}}}
    \item \textcolor{cvprblue}{\textbf{H~~Entropy-Aware Vulnerability Search}} \dotfill {\color{black}\textbf{\pageref{app:entropy_aware_search}}}
    \item \textcolor{cvprblue}{\textbf{I~~Occupancy Map Estimation}} \dotfill {\color{black}\textbf{\pageref{app:bsc_attack}}}
    \item \textcolor{cvprblue}{\textbf{J~~More Background and Related Work}} \dotfill {\color{black}\textbf{\pageref{app:background_related_work}}}
    \begin{itemize}[leftmargin=0.5cm, label={}]
        \item \textcolor{cvprblue}{J.1~~Robust Collaborative Perception} \dotfill {\color{black}\pageref{app:background_related_work}}
        \item \textcolor{cvprblue}{J.2~~Attacks on Collaborative Perception} \dotfill {\color{black}\pageref{app:background_related_work}}
        \item \textcolor{cvprblue}{J.3~~Defensive Collaborative Perception} \dotfill {\color{black}\pageref{app:background_related_work}}
    \end{itemize}
    \item \textcolor{cvprblue}{\textbf{K~~Structure of MVIGNet}} \dotfill {\color{black}\textbf{\pageref{app:mvig_structure}}}
\end{itemize}

\vspace{0.5em}

\section{Analysis of MVIG Representation}
\label{app:analysis_mvig_representation}

In this section, we provide theoretical analysis of why the unified MVIG representation can effectively reveal vulnerabilities exposed by different CP defense systems.

\noindent \textbf{Overview.} Our analysis consists of two complementary parts. Section \ref{subsec:info_theoretic} establishes the \textit{system-level} foundation through spectral graph analysis, proving that vulnerabilities manifest as specific spectral signatures in MVIG. Section \ref{subsec:entropy_based} develops a \textit{region-level} quantification framework using entropy analysis, enabling precise vulnerability measurement at individual spatial locations and adaptation to different defense mechanisms.

\subsection{System-Level Vulnerability Characterization}
\label{subsec:info_theoretic}

To understand why MVIG can effectively detect vulnerabilities across different CP defenses, we establish a unified spectral framework. This framework combines information theory with graph spectral analysis to prove that vulnerabilities manifest as specific spectral signatures—information flow capacity and consensus fragility—that can be systematically identified through MVIG's structure.

\noindent \textbf{Unified Information-Spectral Framework.} Let $O_i$ denote the occupancy matrix observed by CAV $i$. Following the MVIG construction in the main text, the mutual information between CAVs $i$ and $j$ is:
\begin{equation}
\mathcal{I}(O_i; O_j) = \sum_{a,b=0}^{2} p_{ij}(a,b) \log\frac{p_{ij}(a,b)}{p_i(a)p_j(b)},
\end{equation}
where $p_{ij}(a,b)$ is the joint probability and $p_i(a), p_j(b)$ are marginal probabilities. The MVIG edge weights aggregate this across all spatial positions:
\begin{equation}
\mathbf{W}_{ij} = \mathbb{E}_{(x,y)}[\mathcal{I}(O_i; O_j)].
\end{equation}
The spectral decomposition $\mathbf{W} = \mathbf{Q}\mathbf{\Lambda}\mathbf{Q}^T$ reveals the vulnerability structure. We define the \textit{information flow capacity} as:
\begin{equation}
C_{flow} = \frac{\lambda_1}{n-1},
\end{equation}
and the \textit{consensus fragility} as:
\begin{equation}
F_{frag} = \frac{1}{\lambda_2 + \epsilon},
\end{equation}
where $\lambda_1$ is the largest eigenvalue of $\mathbf{W}$ and $\lambda_2$ is the second smallest eigenvalue of the Laplacian $\mathbf{L} = \mathbf{D} - \mathbf{W}$.

\noindent \textbf{Vulnerability Characterization Theorem.} We establish the fundamental relationship between information asymmetry and attack success:

\noindent \textit{Theorem 1 (Vulnerability-Spectral Correspondence):} A CP system is vulnerable to fabrication attacks if and only if $C_{flow} < \tau_1$ and $F_{frag} > \tau_2$ for system-dependent thresholds $\tau_1, \tau_2$.

\noindent \textit{Proof:} (\textit{Necessity}) If a CP system can be successfully attacked, then there must exist information gaps among CAVs (manifested as low $C_{flow}$) and weak consensus validation mechanisms (manifested as high $F_{frag}$). Without information gaps, all CAVs share complete knowledge and can detect fabrications through perfect cross-validation. Without weak consensus, the system can reliably identify inconsistent observations.

(\textit{Sufficiency}) We construct an attack strategy based on spectral analysis. When $C_{flow}$ is low, the principal eigenvector of $\mathbf{W}$ identifies CAVs with weak information coupling, indicating regions where different CAVs have asymmetric views. When $F_{frag}$ is high, the Fiedler vector (eigenvector of the second smallest eigenvalue of Laplacian $\mathbf{L}$) reveals the optimal bipartition of the CAV network, identifying which CAVs can be isolated or misled. By targeting these vulnerable configurations, we can construct fabrications that exploit the identified spectral weaknesses.

The attack success probability can be modeled as:
\begin{equation}
P_{success} = 1 - P_{detection} \approx 1 - \exp\left(-\alpha F_{frag} (1 - C_{flow})\right),
\end{equation}
where $\alpha > 0$ is a system-dependent constant. This exponential form captures the intuition that detection probability decays exponentially as vulnerabilities increase: higher consensus fragility $F_{frag}$ weakens collective validation, while lower information flow capacity $C_{flow}$ (i.e., higher $1-C_{flow}$) indicates greater information asymmetry. The exponential relationship models the multiplicative effect where these two factors jointly determine system vulnerability. Note that $C_{flow} \in [0, 1]$ by construction as a normalized metric. \hfill $\square$

\subsection{Region-Level Vulnerability Quantification}
\label{subsec:entropy_based}

Theorem 1 identifies system-level vulnerability conditions but cannot pinpoint vulnerable regions at specific spatial locations. To enable practical vulnerability assessment, we introduce an entropy-based quantification framework. This framework measures local vulnerabilities through the \textit{entropy deficit} between collective consensus and individual observations, and we prove its effectiveness across different defense mechanisms.

\noindent \textbf{Unified Entropy-Deficit Framework.} For a spatial region $R$, we define the \textit{information entropy deficit} as the fundamental vulnerability measure. The collective uncertainty is:
\begin{equation}
H_c(R) = -\sum_{s=0}^{2} p_c(s|R) \log p_c(s|R),
\end{equation}
where $s \in \{0,1,2\}$ represents occupancy states (free, occupied, unknown) and $p_c(s|R)$ is the consensus probability. The individual entropy of CAV $i$ is:
\begin{equation}
H_i(R) = -\sum_{s=0}^{2} p_i(s|R) \log p_i(s|R).
\end{equation}
The vulnerability score quantifies the entropy deficit:
\begin{equation}
V(R) = H_c(R) - \frac{1}{n} \sum_{i=1}^{n} H_i(R),
\end{equation}
which captures the fundamental trade-off between collective uncertainty and individual confidence. When $V(R) > 0$, the collective consensus exhibits higher uncertainty than individual observations, indicating vulnerable regions where attackers can exploit disagreements among CAVs. Conversely, when $V(R) \leq 0$, individual CAVs have high uncertainty while the collective consensus is more certain, typically occurring in regions with sufficient multi-view coverage where attacks are less effective.

\noindent \textbf{Theoretical Analysis and Bounds.} We establish rigorous theoretical guarantees for vulnerability detection:

\noindent \textit{Theorem 2 (Entropy-Deficit Vulnerability Theorem):} A region $R$ is vulnerable to fabrication attacks if and only if $V(R) > \tau$ for some threshold $\tau > 0$. Moreover, the attack success probability satisfies:
\begin{equation}
P_{attack}(R) \geq \sigma(\beta V(R) - \tau),
\end{equation}
where $\sigma(\cdot)$ is the sigmoid function and $\beta > 0$ is a scaling parameter. The sigmoid function is chosen to provide smooth, bounded probability estimates in $[0,1]$ and ensure differentiability for gradient-based optimization. The threshold $\tau$ is empirically determined based on the defense system's validation mechanism and can be estimated from training data by analyzing the entropy deficit distribution in successful attack regions.

\noindent \textit{Proof:} The necessity follows from information-theoretic arguments: successful attacks require exploiting regions where collective consensus is weak relative to individual confidence. For sufficiency, we construct an explicit attack that targets the entropy deficit. The bound follows from the relationship between entropy deficit and the probability of consensus failure, where the sigmoid function models the smooth transition between secure and vulnerable regions as the entropy deficit increases. \hfill $\square$

\noindent \textbf{Defense-Adaptive Information Aggregation.} The MVIG framework adapts to different defense mechanisms through information-theoretic principles:

For occupancy-based defenses (e.g., CAD), the mutual information computation becomes:
\begin{equation}
\mathcal{I}_{CAD}(O_i; O_j) = \sum_{(x,y) \in R} \mathcal{I}(O_i(x,y); O_j(x,y)),
\end{equation}
providing fine-grained spatial analysis. For feature-based defenses, we employ Blind Region Segmentation with bounded estimation error:
\begin{equation}
\mathcal{I}_{BRS}(\hat{O}_i; \hat{O}_j) = \mathcal{I}(\hat{O}_i; \hat{O}_j) + \epsilon,
\end{equation}
where $|\epsilon| \leq \delta$ and $\delta$ depends on the BRS algorithm accuracy.

\noindent \textit{Corollary 1 (Defense-Invariant Vulnerability):} Despite estimation errors, the vulnerability detection remains robust: if $V_{true}(R) > \tau + 2\delta$, then $V_{est}(R) > \tau$ with probability at least $1 - \exp(-\gamma \delta^2)$ for some $\gamma > 0$.



\section{Loss Function}
\label{app:loss_functions}

\subsection{MVIGNet Loss Function}

We train the MVIGNet using a multi-objective loss function that evaluates the effectiveness of the mask positions:
\begin{equation}\label{eq:total_loss}
    \mathcal{L}(\theta; \mathcal{D}) = \alpha\mathcal{L}_{\text{a}}(\theta; \mathcal{D}) + \beta\mathcal{L}_{\text{b}}(\theta; \mathcal{D}) + \gamma\mathcal{L}_{\text{d}}(\theta; \mathcal{D}),
\end{equation}
where $\theta$ represents MVIGNet parameters, $\mathcal{D}$ is the training dataset, and $\alpha, \beta, \gamma \in \mathbb{R}^+$ are balancing coefficients. The three loss components address distinct aspects of attack optimization:

\noindent \textbf{Attack Effectiveness Loss.}
Let $\mathcal{S}$ and $\mathcal{R}$ denote spoofing and removal attacks respectively. The attack effectiveness loss $\mathcal{L}_{\text{a}}$ is defined as:
For spoofing attacks ($\mathcal{A} = \mathcal{S}$), the attack effectiveness loss maximizes detection confidence:
\begin{equation}
    \mathcal{L}_{\text{a}}(\mathcal{S}) = \sum_{b \in B'} \text{IoU}(b, b_t) \cdot \log(1 - b_{\sigma}) + \lambda_d d(b, v),
\end{equation}
For removal attacks ($\mathcal{A} = \mathcal{R}$), the attack effectiveness loss minimizes detection confidence:
\begin{equation}
    \mathcal{L}_{\text{a}}(\mathcal{R}) = -\sum_{b \in B'} \text{IoU}(b, b_t) \cdot \log(1 - b_{\sigma}) + \lambda_d d(b, v),
\end{equation}
where $B'$ denotes the set of bounding box proposals after applying the perturbation, $b_{\sigma}$ is the confidence score associated with proposal $b$, $b_t$ represents the target region to attack, and $d(b, v)$ is the spatial distance between the proposed box and the victim CAV position with $\lambda_d$ as a weighting factor. This objective function either maximizes or minimizes the confidence scores of proposals overlapping with the target region, depending on whether we're performing a spoofing or removal attack, while penalizing attacks that are too distant from the victim CAV to ensure attacks occur within effective perception range.

\noindent \textbf{Box Differentiation Loss.}
To ensure that spoofed objects appear distinct from existing objects, we apply a box differentiation loss:
\begin{equation}
    \mathcal{L}_{\text{b}}(\mathcal{A}) = 
    \begin{cases}
        \max_{b_i \in B} \text{IoU}(b_t, b_i), & \mathcal{A} = \mathcal{S} \\
        0, & \mathcal{A} = \mathcal{R}
    \end{cases}
\end{equation}
where $B$ represents the set of all detected bounding boxes. This loss penalizes spoofed objects that overlap significantly with existing objects, encouraging the creation of distinct new objects rather than modifications to existing ones. For removal attacks, this loss is not applicable as the goal is to remove existing objects rather than create new ones.

\noindent \textbf{Defense Evasion Loss.}
To evade defense mechanisms that rely on occupancy map cross-verification, we implement a defense-aware loss component:
\begin{equation}
    \mathcal{L}_{\text{d}} = \sigma(\eta(\tau - d_{\min}(E_{\text{conflict}}, b_t)))
\end{equation}
where $\sigma$ is the sigmoid function ensuring smooth gradients, $\eta$ is a scaling factor controlling the steepness of the penalty, $\tau$ is a safety threshold distance, $d_{\min}(E_{\text{conflict}}, b_t) = \min_{e \in E_{\text{conflict}}} \|e - b_t\|_2$ computes the minimum Euclidean distance between the target box center $b_t$ and any conflict region $e$, and $E_{\text{conflict}}$ represents the identified conflict areas. This loss encourages attacks to maintain sufficient distance from regions that would trigger defense systems' inconsistency detection.

\begin{figure*}[t]
  \centering
  \includegraphics[width=\linewidth]{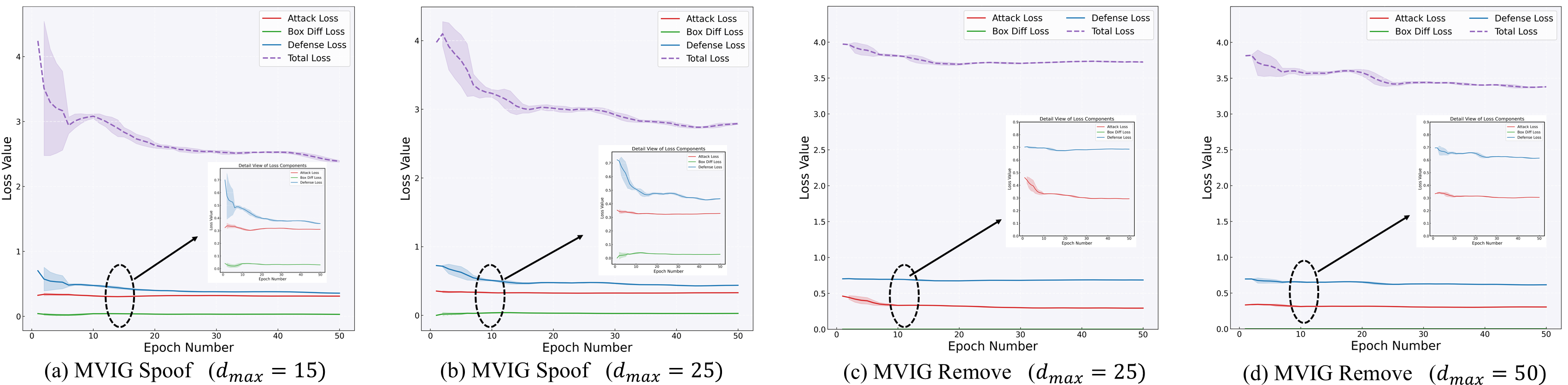}
  \caption{Validation loss curves of MVIGNet training on OPV2V dataset.}
  \label{fig:model_convergence}
\end{figure*}

\noindent \textbf{Visualization of MVIGNet Loss Curves.}
Figure \ref{fig:model_convergence} shows the validation loss curves of different components during MVIGNet training. For the MVIG-Spoof attack (left), we observe that the attack loss stabilizes around 0.35 after approximately 20 epochs, while defense loss rapidly decreases to near 0.05, indicating effective evasion of defense mechanisms. The box difference loss remains consistently low throughout training, ensuring spatial accuracy of injected objects. In contrast, the MVIG-Removal attack (right) exhibits a different pattern where defense loss dominates (around 0.2), suggesting that evading detection of removed objects is more challenging than spoofing new ones. The attack loss converges to a lower value (approximately 0.07), demonstrating that the model efficiently learns to remove existing objects. In both attack types, the total loss shows initial volatility but stabilizes after 20 epochs, with the removal attack requiring slightly more training iterations to converge. The shaded regions represent confidence intervals across multiple training runs, confirming the consistency and reproducibility of our approach.

\subsection{PGD Loss Function}
The three loss components described above ($\mathcal{L}_{\text{a}}$, $\mathcal{L}_{\text{b}}$, and $\mathcal{L}_{\text{d}}$) train our MVIGNet to predict optimal mask locations. While PGD optimization uses a loss function functionally similar to $\mathcal{L}_{\text{a}}$, the two procedures operate independently.
Once MVIGNet identifies the optimal attack mask $M_t^*$, we generate the feature perturbation through two steps. First, we initialize it based on the predicted mask:
\begin{equation}
    \delta_t^{(0)} = \Phi_{\psi}(M_t^*),
\end{equation}
where $\Phi_{\psi}: \{0,1\}^{H \times W} \rightarrow \mathbb{R}^d$ maps the binary mask to feature space. Then, we apply PGD to refine the perturbation:
\begin{equation}
    \delta_t^{(k+1)} = \Pi_{\Omega_{M_t^*}}(\delta_t^{(k)} - \alpha \cdot \nabla_{\delta_t^{(k)}}\phi(\delta_t^{(k)})),
\end{equation}
where $\alpha$ is the step size, $\nabla_{\delta_t^{(k)}}\phi(\delta_t^{(k)})$ is the gradient of PGD loss, and $\Pi_{\Omega_{M_t^*}}$ projects the perturbation onto the feasible region. Note that while PGD typically performs gradient ascent for maximization, we use gradient descent here because the PGD loss $\phi$ is formulated as a minimization objective that aligns with $\mathcal{L}_{\text{a}}$, i.e., minimizing $\phi$ corresponds to maximizing detection confidence for spoofing or minimizing it for removal attacks.
The final perturbation $\delta_t$ is added to the original feature map $\mathbf{F}_a^t$ to produce the manipulated feature $\mathbf{F}_a^t + \boldsymbol{\delta}$ sent to victim CAVs.

\section{More Implementation Details}
\label{app:more_implementation_details}

\subsection{Evaluation Metrics}
\label{app:evaluation_metrics}

\noindent 	\textbf{Attack Success Rate (ASR)}: This metric quantifies the effectiveness of an attack in a scenario without any defense mechanism in place. It is calculated as the percentage of attack attempts that successfully mislead the ego vehicle's perception, for instance, by causing it to detect a non-existent object (spoofing) or miss a real one (removal). A higher ASR indicates a more potent attack.

\noindent \textbf{Defense Success Rate (DSR)}: Conversely, in scenarios where a defense mechanism is active, DSR measures the percentage of incoming attacks that are correctly identified and neutralized by the defender. From the attacker's perspective, a lower DSR is desirable as it signifies that the attack is more successful at evading detection.

\noindent \textbf{$\Delta$AP@50}: This metric measures the degradation in the victim's object detection \citep{guo2025neptunexactivextomaritimegeneration} performance caused by an attack. AP (Average Precision) is a standard metric for detection accuracy, and the IoU (Intersection over Union) threshold is set to 50\%. $\Delta$AP@50 represents the change (typically a drop) in AP, providing a clear measure of the attack's impact on the perception system's reliability.

\noindent \textbf{True Positive Rate (TPR)}: Also known as sensitivity or recall, TPR measures the proportion of actual malicious data transmissions that are correctly flagged as attacks by the defense system. A high TPR is crucial for a robust defense.

\noindent \textbf{False Positive Rate (FPR)}: This measures the proportion of legitimate, benign data transmissions that are incorrectly identified as malicious by the defense system. A high FPR is undesirable as it leads to the unnecessary rejection of valid data, which can degrade the overall performance and benefits of collaborative perception.

\noindent \textbf{Frames Per Second (FPS)}: This metric indicates the computational efficiency of an attack or defense algorithm by measuring how many frames of data it can process per second. For real-world autonomous driving applications, a high FPS is essential to ensure that the system can operate in real-time without introducing dangerous latency.

\subsection{Baselines}
\label{app:baselines}

\noindent \textbf{Basic Feature Attack} \cite{9711249}: This is a foundational, untargeted attack method that directly perturbs the feature maps shared between vehicles. It acts as a baseline for adversarial capability but lacks sophistication, as the perturbations are not optimized for stealth or for specific regions, making it relatively easy to detect.

\noindent \textbf{Blind Area Confusion (BAC) Attack} \cite{tao2025gcpguardedcollaborativeperception}: A more advanced untargeted attack, BAC leverages knowledge of the victim's field of view to constrain its perturbations primarily to the victim's blind spots. This spatial constraint increases the attack's stealthiness compared to the basic feature attack.

\noindent \textbf{Ray-Casting (RC) Attack} \cite{294490}: A targeted fabrication attack that aims to create a specific "ghost" vehicle at a chosen location. While it can generate subtle and effective perturbations, its primary limitation is the lack of a strategic mechanism to optimize the attack's timing and target region, which our \texttt{MVIG} attack is designed to overcome.

\noindent \textbf{CAD (Collaborative Anomaly Detection)} \cite{294490}: A defense system that establishes a consensus by comparing the occupancy maps from different collaborating vehicles. Anomalies, indicated by significant disagreements between maps, are flagged as potential attacks. Its effectiveness relies on having sufficient overlapping views.

\noindent \textbf{ROBOSAC} \cite{Li_2023_ICCV}: A defense framework built on the robust RANSAC (Random Sample Consensus) algorithm, which treats malicious data as outliers to a consensus model derived from a random subset of collaborators. The enhanced variant, ROBOSAC+, incorporates temporal verification, checking for object consistency across consecutive frames to counter transient attacks.

\noindent \textbf{CP-Guard} \cite{hu2024cpguardmaliciousagentdetection}: This defense system uses PASAC (Probability-Agnostic Sample Consensus), an improvement over traditional RANSAC that does not require prior assumptions about the distribution of malicious agents. This makes it more adaptive and robust in scenarios with varying numbers of attackers.

\noindent \textbf{GCP (Guarded Collaborative Perception)} \cite{tao2025gcpguardedcollaborativeperception}: A sophisticated defense mechanism that performs rigorous spatial-temporal consistency checks across multiple frames. It verifies the plausibility of detected objects by analyzing their position and movement over time, making it challenging for fabricated objects that lack realistic trajectories to remain undetected.

\section{More Experimental Results}
\label{app:more_experimental_results}

\subsection{Visualization of Feature Maps}
\begin{figure*}[t]
    \centering
    \includegraphics[width=\linewidth]{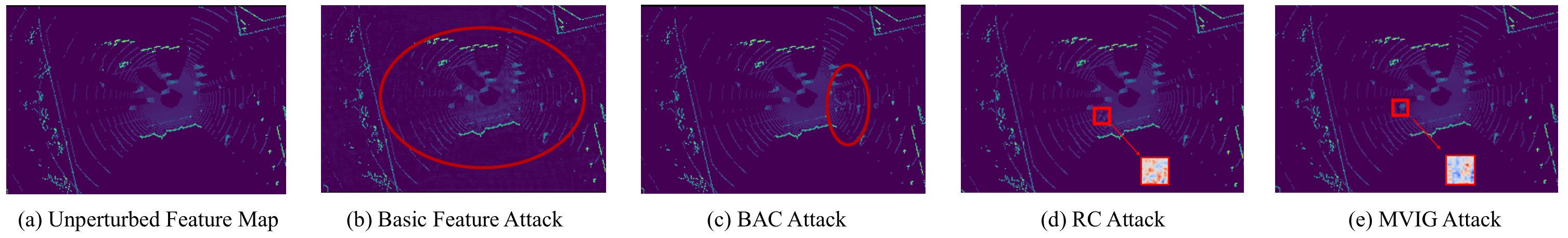}
    \caption{Feature map comparison of different attack methods on Adv-OPV2V dataset.}
    \label{fig:feature_maps}
\end{figure*}

Figure \ref{fig:feature_maps} visualizes feature maps under different attacks. Basic feature attack and BAC attack generate untargeted perturbations covering widespread areas, making them easily detectable. In contrast, RC attack and \texttt{MVIG} attack focus perturbations within specific target regions. \texttt{MVIG} attack tends to identify and exploit LiDAR-sparse regions near obstacles for spoofing, as these locations are likely mutual blind spots to all benign CAVs. 

\subsection{Results with Different CP Architectures}
\label{app:cp_architectures}

To evaluate the generalizability of our \texttt{MVIG} attack across different CP architectures, we conducted additional experiments on mainstream CP models including AttFuse~\cite{294490}, V2VNet~\cite{10.1007/978-3-030-58536-5_36}, and Where2comm~\cite{huWhere2commCommunicationefficientCollaborative2024}. Table \ref{tab:cp_architectures} shows the defense success rates of different defenders against MVIG spoof attack under various CP architectures.
\begin{table}[t]
  \centering
  \caption{Defense success rates (\%) across different CP architectures. Lower values indicate better attack effectiveness.}
  \label{tab:cp_architectures}
  \small
  \setlength{\tabcolsep}{2pt}
  \renewcommand{\arraystretch}{1.2}
  \begin{tabular}{l|c|c|c|c}
  \hline
  \textbf{CP Architecture} & \textbf{ROBOSAC} & \textbf{CP-Guard} & \textbf{GCP} & \textbf{CAD} \\
  \hline
  AttFuse (Full) & 14.8 & 17.2 & 13.0 & 32.0 \\
  V2VNet (Full) & 14.9 & 17.1 & 13.5 & 32.2 \\
  Where2comm (Sparse) & 15.2 & 18.1 & 13.6 & 33.1 \\
  \hline
  \end{tabular}
\end{table}
The results demonstrate that our \texttt{MVIG} attack maintains consistent effectiveness across different CP architectures, with only minor variations in performance. This indicates that our attack framework is not sensitive to specific architectural choices and can generalize well to various feature fusion strategies, including both full and sparse feature transmission methods.

\subsection{Scalability Analysis}
\label{app:scalability}

We evaluated the scalability of our \texttt{MVIG} attack by varying the number of benign CAVs in the CP network. Table \ref{tab:scalability} shows the defense success rates against MVIG spoof attack as the number of participating benign CAVs increases.
\begin{table}[t]
  \centering
  \caption{Defense success rates (\%) with varying numbers of benign CAVs. Lower values indicate better attack effectiveness.}
  \label{tab:scalability}
  \small
  \setlength{\tabcolsep}{3.5pt}
  \renewcommand{\arraystretch}{1.2}
  \begin{tabular}{c|c|c|c|c}
  \hline
  \textbf{No. Benign CAVs} & \textbf{ROBOSAC} & \textbf{CP-Guard} & \textbf{GCP} & \textbf{CAD} \\
  \hline
  1 & 13.6 & 16.4 & 12.5 & 15.1 \\
  2 & 14.8 & 17.2 & 13.0 & 32.0 \\
  3 & 14.9 & 18.0 & 13.2 & 36.2 \\
  \hline
  \end{tabular}
\end{table}
As expected, we observe a slight degradation in \texttt{MVIG} attack performance as the number of benign CAVs increases, particularly against the CAD defender which benefits from more comprehensive occupancy map validation. However, our \texttt{MVIG} attack still maintains good effectiveness (less than 37\% defense success rate against CAD) within current dataset limitations, where most existing CP datasets contain at most 2-5 connected agents.

\subsection{Robustness to Occupancy Map  Errors}
\label{app:robustness}

We conducted experiments to evaluate \texttt{MVIG}'s sensitivity to occupancy map corruption caused by blind region segmentation (BRS) estimation errors. This is particularly relevant when attackers must estimate occupancy information from feature maps rather than accessing precise occupancy maps directly. Table \ref{tab:robustness} shows the defense success rates under different BRS error rates.
\begin{table}[t]
  \centering
  \caption{Defense success rates (\%) under different BRS error rates. Lower values indicate better attack effectiveness.}
  \label{tab:robustness}
  \small
  \setlength{\tabcolsep}{4pt}
  \renewcommand{\arraystretch}{1.2}
  \begin{tabular}{c|c|c|c|c}
  \hline
  \textbf{BRS Error Rate} & \textbf{ROBOSAC} & \textbf{CP-Guard} & \textbf{GCP} & \textbf{CAD} \\
  \hline
  10\% & 14.8 & 17.0 & 13.0 & 34.2 \\
  20\% & 15.1 & 17.4 & 13.1 & 40.5 \\
  30\% & 15.2 & 17.3 & 13.2 & 52.1 \\
  \hline
  \end{tabular}
\end{table}
The results show that our \texttt{MVIG} attack is not sensitive to occupancy map corruption caused by BRS estimation errors for most defenders (ROBOSAC, CP-Guard, GCP). This robustness stems from the fact that these defenders rely on holistic detection map similarity rather than fine-grained occupancy validation. For CAD defense, which requires precise occupancy maps, our attack can still tolerate estimation error rates up to 30\% while maintaining reasonable effectiveness. However, it is worth noting that CAD defense inherently requires CAVs to transmit precise occupancy maps for validation, meaning attackers would not encounter such high estimation errors in practice. Therefore, our \texttt{MVIG} attack remains robust against all current CP defensive systems.

\section{More Discussions}
\label{app:more_discussions}

\subsection{Attack Effectiveness Analysis}
\label{app:attack_effectiveness}

We provide deeper analysis of why MVIG spoof attacks demonstrate superior performance compared to other attack methods and removal attacks:

\noindent \textbf{MVIG vs. Baseline Attacks:} MVIG's superior performance against defenses stems from its strategic utilization of implicit view confidence information in CP messages to target mutual vulnerable regions of multiple CAVs. Unlike baseline attacks that act as "blind strikes" without considering when and where to attack, MVIG operates like a "strategic sniper" that first identifies shared uncertain areas among neighboring benign CAVs through feature map or occupancy data analysis, then strategically guides perturbations towards these mutual vulnerabilities.

\noindent \textbf{MVIG Spoof vs. Removal:} MVIG spoof attacks show better performance than removal attacks due to larger optimization search space of candidate fabrication regions. Removal attacks must target existing objects, making the candidate fabrication region quite limited, while spoof attacks can explore broader areas for object placement.

\noindent \textbf{CAD Defense Analysis:} CAD shows better defense performance because it performs fine-grained grid-by-grid validation using additional precise occupancy maps that are required to be exchanged among CAVs. In contrast, other defenses like ROBOSAC and CP-Guard do not require exchanging such validation data and only rely on comparing holistic detection map similarity between neighboring CAVs and ego CAV to identify anomalies.

\subsection{Potential Countermeasures Discussion}
\label{app:countermeasures}

Based on our analysis of \texttt{MVIG} attack mechanisms, we identify several potential countermeasure directions:

\noindent \textbf{Confidence Pattern Obscuring:} A potential defense could involve obscuring confidence patterns in transmitted data using differential privacy-based methods to inject calibrated noise on feature maps or occupancy maps, or employing confidence quantization techniques to disrupt MVIG's ability to identify mutual vulnerable regions.

\noindent \textbf{Randomized View Sharing:} Implementing randomized or encrypted view sharing protocols could make it more difficult for attackers to consistently analyze mutual view information across multiple frames.

\noindent \textbf{Temporal Anomaly Detection:} Enhanced temporal consistency checks that monitor for sudden changes in collaboration patterns or unexpected persistent objects could help detect \texttt{MVIG} attacks.

However, these countermeasures create a fundamental trade-off between collaborative perception accuracy and attack resilience, representing an interesting direction for future research in secure collaborative perception systems.

\subsection{Communication Delay}
\label{app:communication_delay}

In our work, we follow the standard assumption adopted by prior CP security literature \citep{Li_2023_ICCV,294490,tao2025gcpguardedcollaborativeperception} by focusing on low-latency CP systems where feature map transmission and caching can be accomplished within one frame (typically less than 100ms). An ego CAV collects feature maps from neighboring CAVs based on timestamps that fall into a certain frame time window. 

For high-latency scenarios, prior works focusing on communication issues in CP systems \citep{lei2022latencyawarecollaborativeperception,NEURIPS2023_5a829e29} have demonstrated that latency can be addressed through feature prediction and compensation techniques. Our \texttt{MVIG} attack can be readily adapted to such high-latency scenarios through two approaches: (1) incorporating feature prediction and compensation into our attack optimization process, or (2) strategically waiting for better channel conditions to initiate attacks when the latency is reduced.

\noindent \textbf{Synchronization Considerations.} In our threat model, we consider an internal attacker scenario where the attacker has successfully compromised a legitimate CAV's data access rights. In this scenario, the attacker can normally interact with neighboring benign CAVs to exchange feature maps and occupancy maps (if shared) until it is identified as a malicious agent by defenders. The attacker leverages the same synchronization mechanisms as legitimate CAVs, accessing the same timestamp-based frame windows used by the CP system. This eliminates the need for additional synchronization capabilities beyond what the CP system already provides.

\noindent \textbf{Real-World Deployment.} Modern V2X communication systems typically achieve latencies of 20-100ms under normal conditions, which aligns well with our assumption. For scenarios where network conditions degrade, the attacker can employ adaptive strategies such as monitoring channel quality and postponing attacks until conditions improve, or adjusting the prediction horizon $m$ to accommodate longer transmission delays. These practical considerations do not fundamentally change our attack framework but rather represent operational parameters that can be tuned based on deployment conditions.

\section{Overhead Analysis}
\label{app:overhead_analysis}

We analyze the computational overhead of our \texttt{MVIG} attack and compare it with baseline methods. All experiments are conducted on a server with Intel Xeon Silver 4410Y CPU and NVIDIA RTX A5000 GPU.

\subsection{Runtime Performance}
Table \ref{tab:runtime} shows the frame rates (FPS) achieved by different attack methods under various persistence settings. The persistence parameter $p$ represents the number of consecutive frames covered by a single optimization, with higher values indicating more efficient resource utilization. Our \texttt{MVIG} attack achieves 15.6-29.9 FPS, demonstrating real-time feasibility for practical deployment.

\begin{table}[t]
  \centering
  \caption{Runtime comparison across different attack methods (FPS).}
  \label{tab:runtime}
  \small
  \setlength{\tabcolsep}{10pt}
  \renewcommand{\arraystretch}{1.2}
  \begin{tabular}{@{}l|c|c|c@{}}
  \hline
  \textbf{Methods} & \textbf{0-frame} & \textbf{1-frame} & \textbf{3-frame} \\
  \hline
  Basic Attack \cite{9711249} & 24.4 & 24.4 & 24.4 \\
  RC Attack \cite{294490} & 17.9 & 24.2 & 33.8 \\
  BAC Attack \cite{tao2025gcpguardedcollaborativeperception} & 10.8 & 15.0 & 18.4 \\
  \texttt{MVIG} Attack (Ours) & 15.6 & 21.3 & 29.9 \\
  \hline
  \end{tabular}
\end{table}

\subsection{Resource Requirements}
Our \texttt{MVIG} attack requires modest computational resources:
\begin{itemize}
    \item \textbf{GPU memory}: ~5.6 GB (2.6 GB for mask generation + 3.0 GB for PGD optimization)
    \item \textbf{Processing time}: 115 ms total (14 ms MVIGNet + 88 ms PGD + 13 ms others)
    \item \textbf{CPU operations}: Feature mapping and coordinate transformations
\end{itemize}

This makes our attack feasible on automotive-grade hardware like NVIDIA DRIVE AGX Thor (250 TFLOPS, 16GB memory) and even mainstream edge devices like Jetson Orin NX (3.8 TFLOPS, 16GB memory).

\subsection{Prediction Horizon Selection}
To ensure online attack feasibility, we must complete all preparation work before the target frame arrives. As shown in Table \ref{tab:attack_runtime}, a single \texttt{MVIG} attack instance comprises four operations: mask generation, PGD optimization, attack transformation, and feature mapping. For the OPV2V dataset operating at 10 FPS (100 ms per frame), setting the prediction horizon to $m = 2$ provides a 200 ms time buffer, which is sufficient to complete all attack operations before the target frame arrives. Specifically, with $m = 2$:

\begin{enumerate}
    \item The MVIGNet processes the current frame and predicts attack positions in 14 ms.
    \item PGD optimization for initial mask generation requires 88 ms.
    \item Attack transformation takes 3 ms.
    \item Other operations (feature mapping, coordinate transformation, etc.) require 10 ms.
\end{enumerate}

The total processing time of 115 ms is well within the 200 ms time window provided by a 2-frame prediction horizon, ensuring our attack can operate in real-time while maintaining effectiveness. Larger values of $m$ would increase prediction uncertainty without providing significant timing benefits, while $m = 1$ would be insufficient to complete all operations before the target frame arrives.

\subsection{Component-wise Analysis}
Table \ref{tab:attack_runtime} provides detailed breakdown of computational costs for different attack components across methods.

\begin{table*}[t]
    \centering
    \small
    \caption{Runtime comparison of different attack methods. The persistence parameter $p$ represents the number of consecutive frames covered by a single optimization. N/A indicates the process is not part of the attack, while Offline indicates the process is pre-computed and not included in real-time application.}
    \label{tab:attack_runtime}
    \renewcommand{\arraystretch}{1.2}
    \setlength{\tabcolsep}{8.5pt}
    \begin{tabular}{l|ccccc|ccc}
    \hline
    \multirow{2}{*}{\textbf{Method}} & \multicolumn{5}{c|}{\textbf{Component Runtime}} & \multicolumn{3}{c}{\textbf{Avg. FPS}} \\
    \cline{2-9}
    & Mask Gen. & Ray-Cast & PGD Opt. & Attack Trans. & Others & $p=0$ & $p=1$ & $p=3$ \\
    \hline
    Basic Attack \cite{9711249} & N/A & N/A & 40 ms & N/A & 1 ms& 24.4 & 24.4 & 24.4 \\
    BAC Attack \cite{tao2025gcpguardedcollaborativeperception} & 49/(1+$p$) ms & N/A & 42 ms & N/A & 2 ms& 10.8 & 15.0 & 18.4 \\
    RC Attack \cite{294490} & N/A & Offline & 88/(2+$p$) ms & 3 ms & 9 ms & 17.9 & 24.2 & 33.8 \\
    \texttt{MVIG} Attack (Ours) & 14/(2+$p$) ms & Offline & 88/(2+$p$) ms & 3 ms & 10 ms & 15.6 & 21.3 & 29.9 \\
    \hline
    \end{tabular}
\end{table*}

\section{Fabrication Risk Map Generation}
\label{app:attack_risk_map}

During inference, we employ Gaussian kernel-based contrast enhancement as test-time augmentation to improve attack position decision robustness. This process refines the masked score map $\hat{\mathbf{S}}_{t+m} \in \mathbb{R}^{H \times W}$ into a risk representation $\tilde{\mathbf{S}}_{t+m} \in [0,1]^{H \times W}$ that highlights vulnerability regions while suppressing noise.
A simple normalization approach would define the risk map as:
\begin{equation}
    \mathbf{R}_{d} = \frac{\hat{\mathbf{S}}_{t+m}}{\|\hat{\mathbf{S}}_{t+m}\|_{\infty}},
\end{equation}
where $\|\cdot\|_{\infty}$ denotes the maximum norm. However, this elementary transformation fails to capture intrinsic spatial correlations and vulnerability variations, potentially leading to inconsistent attack decisions.
Our enhanced processing pipeline consists of three steps. First, we apply Gaussian kernel smoothing:
\begin{align}
    \mathbf{S}_{s} &= (\mathcal{G}_{\sigma} * \hat{\mathbf{S}}_{t+m})(x,y) \nonumber \\
    &= \iint_{\mathbb{R}^2} \hat{\mathbf{S}}_{t+m}(u,v) \cdot G_{\sigma}(x-u, y-v) \,du\,dv,
\end{align}
where the Gaussian kernel $G_{\sigma}(x,y)$ is defined as:
\begin{equation}
    G_{\sigma}(x,y) = \frac{1}{2\pi\sigma^2}\exp\left(-\frac{x^2 + y^2}{2\sigma^2}\right),
\end{equation}
where $\sigma=4.0$ controls the spatial correlation scale. This value corresponds to approximately 2 grid cells ($4.0/2.0$) in our BEV representation (grid resolution $0.4$m), capturing local vulnerability patterns while filtering high-frequency noise from MVIGNet predictions. Next, we perform contrast enhancement:
\begin{equation}
    \mathbf{S}_{c} = \mathcal{T}_{\gamma}(\mathbf{S}_{s}) = \left(\frac{\mathbf{S}_{s}}{\|\mathbf{S}_{s}\|_{\infty}}\right)^{1/\gamma},
\end{equation}
with $\gamma=2.5$. This power-law transformation with $\gamma > 1$ compresses high values and expands low values, enhancing the contrast between vulnerable and safe regions—critical for distinguishing subtle vulnerability gradients in complex urban driving scenarios. Finally, we apply a threshold operation:
\begin{equation}
    \tilde{\mathbf{S}}_{t+m} = \Psi_{\tau}(\mathbf{S}_{c}) = \frac{\max(\mathbf{S}_{c} - \tau, 0)}{\|\max(\mathbf{S}_{c} - \tau, 0)\|_{\infty} + \epsilon},
\end{equation}
where $\tau=0.3$ is the risk threshold and $\epsilon$ is a small constant for numerical stability. This threshold filters out low-confidence regions, ensuring attacks target only high-vulnerability areas where collective perception is genuinely weak. The complete transformation is expressed as:
\begin{equation}
    \tilde{\mathbf{S}}_{t+m} = (\Psi_{\tau} \circ \mathcal{T}_{\gamma} \circ \mathcal{G}_{\sigma})(\hat{\mathbf{S}}_{t+m}).
\end{equation}
By applying this enhancement during inference, we achieve three key properties: (1) spatial coherence through Gaussian diffusion, (2) dynamic range expansion via contrast enhancement, and (3) noise suppression through thresholding. The parameters $(\sigma, \gamma, \tau)$ were optimized to maximize mutual information between original and enhanced maps while minimizing noise entropy, making our attack decisions more reliable in complex urban environments.

\section{Entropy-Aware Vulnerability Search}
\label{app:entropy_aware_search}

To efficiently maintain attack persistence across frames while ensuring temporal consistency, we develop an entropy-aware vulnerability search strategy that identifies optimal attack transformations across spatiotemporal dimensions by maximizing the information entropy deficit between collective consensus and individual CAV observations. Given the current mask position $\mathbf{M}_{t+m+j-1}$ and the fabrication risk map $\tilde{\mathbf{S}}_{t+m}$ at the initial prediction time $t+m$, our transformation operator $\mathcal{T}_j$ projects the mask to positions for each of the $j$ subsequent frames while ensuring temporal consistency.

\noindent \textbf{Entropy-Driven Search Direction.} The fabrication risk map $\tilde{\mathbf{S}}_{t+m}$ is a learned representation of the entropy deficit $V(R)$ quantified in Theorem 2, where higher scores indicate regions with greater collective uncertainty relative to individual confidence. The transformation process begins by identifying the center of mass of the current mask $\mathbf{M}_{t+m+j-1}$, denoted as $(x_c, y_c)$. We compute an entropy-aware search direction by combining two critical vector fields: the current velocity field $\mathbf{v}_{\text{cur}} \in \mathbb{R}^2$ derived from historical positions, and the entropy gradient field $\nabla \tilde{\mathbf{S}}_{t+m}$ that points toward regions of increasing vulnerability (i.e., higher entropy deficit). The ideal search direction $\mathbf{d}_{\text{ideal}} \in \mathbb{R}^2$ is expressed as:
\begin{align}
\mathbf{d}_{\text{ideal}} &= \frac{\alpha \mathbf{v}_{\text{cur}} + (1-\alpha) \nabla \tilde{\mathbf{S}}_{t+m}(x_c,y_c)}{||\alpha \mathbf{v}_{\text{cur}} + (1-\alpha) \nabla \tilde{\mathbf{S}}_{t+m}(x_c,y_c)||},
\end{align}
where the blending coefficient $\alpha \in [0.7, 0.95]$ is adaptively determined based on the angular coherence between velocity and gradient vectors:
\begin{align}
\alpha &= f\left(\cos^{-1}\left(\frac{\mathbf{v}_{\text{cur}} \cdot \nabla \tilde{\mathbf{S}}_{t+m}(x_c,y_c)}{||\mathbf{v}_{\text{cur}}|| \cdot ||\nabla \tilde{\mathbf{S}}_{t+m}(x_c,y_c)||}\right)\right).
\end{align}
The function $f: [0, \pi] \rightarrow [0.7, 0.95]$ maps angular differences to blending coefficients, with smaller angular differences resulting in lower $\alpha$ values (allowing the entropy gradient field to exert greater influence), while larger angles favor preserving motion momentum. The $\text{GradS}$ function samples the fabrication risk map in the forward-facing sector around $(x_c, y_c)$ to approximate the gradient direction that maximizes the entropy deficit, effectively identifying the steepest ascent toward vulnerable regions.

\noindent \textbf{Entropy-Maximizing Position Selection.} Within a trajectory-constrained search space $\Omega_{\mathbf{d}}(x_c, y_c, \delta) \subset \mathbb{R}^2$ defined along the ideal direction, where $\delta \in \mathbb{R}^+$ represents the maximum search radius, we identify the optimal position that maximizes the entropy deficit while respecting physical motion constraints:
\begin{equation}
(x^*, y^*) = \argmax_{(x,y) \in \Omega_{\mathbf{d}}(x_c, y_c, \delta)} \left[ \tilde{\mathbf{S}}_{t+m}(x,y) + \mathcal{R}(x,y) \right],
\end{equation}
where $\tilde{\mathbf{S}}_{t+m}(x,y)$ quantifies the entropy deficit at position $(x,y)$ (i.e., the vulnerability score), and the reward function $\mathcal{R}: \mathbb{R}^2 \times \mathbb{R}^2 \times \mathbb{R}^2 \rightarrow \mathbb{R}$ captures directional coherence, velocity consistency, and historical motion alignment with respect to the ideal direction $\mathbf{d}_{\text{ideal}}$ and current velocity $\mathbf{v}_{\text{cur}}$. To maintain physical realism in autonomous driving scenarios, the reward function penalizes attack transformations that violate motion constraints of typical traffic participants:
\begin{equation}\label{eq:reward_function}
\mathcal{R}(x,y) = \beta_d \mathcal{D}(x,y) + \beta_v \mathcal{V}(x,y) + \beta_h \mathcal{H}(x,y).
\end{equation}

\noindent\textbf{Physical Interpretation.} Each reward component enforces realistic motion constraints typical of urban vehicle traffic:

The directional consistency component $\mathcal{D}$ penalizes abrupt heading changes that would be physically implausible for vehicles. In autonomous driving, vehicles cannot instantaneously change direction; they follow smooth trajectories constrained by steering mechanics and lane geometry:
\begin{equation}
    \mathcal{D}\left((x,y), \mathbf{d}_{\text{ideal}}\right) = 1.0 - \frac{|\angle(\mathbf{d}_{\text{ideal}}, (x-x_c, y-y_c))|}{\theta_{\text{max}}},
\end{equation}
where $\angle(\cdot,\cdot)$ computes the angle between vectors and $\theta_{\text{max}}$ is the maximum allowed angular deviation. This enforces smooth steering behavior, preventing unrealistic sharp turns that would immediately raise suspicion.

The velocity consistency component $\mathcal{V}$ enforces speed coherence across frames, mimicking constant or gradually changing velocities typical of real vehicles rather than erratic speed fluctuations:
\begin{equation}
    \mathcal{V}\left((x,y), (x_c, y_c)\right) = 1.0 - \frac{|d_{\text{ideal}} - \|(x-x_c, y-y_c)\|_2|}{d_{\text{max}}},
\end{equation}
where $d_{\text{ideal}}$ represents the expected displacement based on current velocity (typically $v \cdot \Delta t$, where $\Delta t$ is the inter-frame time) and $d_{\text{max}}$ bounds acceptable acceleration/deceleration ranges. This prevents teleportation-like jumps that violate physics.

The historical motion alignment component $\mathcal{H}$ ensures momentum conservation, rewarding trajectories that continue in directions consistent with recent motion history—reflecting vehicle inertia and driver intent continuity:
\begin{align}
    \mathcal{H}\left((x,y), \mathbf{v}_{\text{cur}}\right) &= \frac{\mathbf{v}_{\text{cur}} \cdot (x-x_c, y-y_c)}{\|\mathbf{v}_{\text{cur}}\|_2 \cdot \|(x-x_c, y-y_c)\|_2},
\end{align}
where $\mathbf{v}_{\text{cur}}$ is the normalized velocity vector from previous frames. High $\mathcal{H}$ values indicate forward motion along the established trajectory, while negative values would suggest unnatural backward motion.

The coefficients $\beta_d, \beta_v, \beta_h \in \mathbb{R}^+$ control the relative importance of each component, balancing physical plausibility with attack effectiveness.
The mask is then translated to this new position while preserving its shape through the $\mathcal{T}_M$ function:
\begin{equation}
\mathbf{M}_{t+m+j} = \mathcal{T}_M(\mathbf{M}_{t+m+j-1}, x^*-x_c, y^*-y_c),
\end{equation}
where $\mathcal{T}_M: \{0,1\}^{H \times W} \times \mathbb{R} \times \mathbb{R} \rightarrow \{0,1\}^{H \times W}$ implements a rigid translation of the binary mask.
For each subsequent frame, we evaluate whether to continue the attack based on the expected risk score in a neighborhood around the current mask center:
\begin{align}
\mathcal{C}_{t+m+j} &= \mathbb{I}\left[\mathbb{E}_{(x,y) \in \mathcal{N}(x_c,y_c)}[\tilde{\mathbf{S}}_{t+m}(x,y)] \geq \eta\right],
\end{align}
where $\mathbb{I}[\cdot]$ is the indicator function, $\mathbb{E}[\cdot]$ denotes the expectation operator, and $\eta$ is the continuation threshold. The final attack mask becomes $\mathbf{M}_{t+m+j}^{\text{final}} = \mathcal{C}_{t+m+j} \cdot \mathbf{M}_{t+m+j}$.
\begin{algorithm}[H]
\caption{Entropy-Aware Vulnerability Search for Attack Transformation}
\begin{algorithmic}[1]
\Require Current mask $\mathbf{M}_{t+m+j-1}$, risk map $\tilde{\mathbf{S}}_{t+m}$, velocity $\mathbf{v}_{\text{cur}}$
\Ensure Transformed mask $\mathbf{M}_{t+m+j}^{\text{final}}$
\State $(x_c, y_c) \gets \text{CoM}(\mathbf{M}_{t+m+j-1})$ \Comment{\texttt{Compute center}}
\State $\nabla \tilde{\mathbf{S}} \gets \text{GradS}(\tilde{\mathbf{S}}_{t+m}, x_c, y_c)$ \Comment{\texttt{Entropy gradient estimation}}
\State $\alpha \gets \text{AdaptiveBlend}(\mathbf{v}_{\text{cur}}, \nabla \tilde{\mathbf{S}})$ \Comment{\texttt{Adaptive blending}}
\State $\mathbf{d}_{\text{ideal}} \gets \text{Normalize}(\alpha\mathbf{v}_{\text{cur}} + (1-\alpha)\nabla \tilde{\mathbf{S}})$ 
\State Calculate reward components $\mathcal{D}, \mathcal{V}, \mathcal{H}$ \Comment{\texttt{Motion coherence terms}}
\State Calculate composite reward $\mathcal{R}$ using Eq.~\ref{eq:reward_function}
\State $(x^*, y^*) \gets \argmax_{(x,y)} \{\tilde{\mathbf{S}}_{t+m}(x,y) + \mathcal{R}(x,y)\}$ \Comment{\texttt{Maximize entropy deficit}} 
\State $\mathbf{M}_{t+m+j} \gets \text{TranslateMask}(\mathbf{M}_{t+m+j-1}, x^*-x_c, y^*-y_c)$
\State $\mathcal{C}_{t+m+j} \gets \mathbb{I}[\mathbb{E}_{(x,y) \in \mathcal{N}(x_c,y_c)}[\tilde{\mathbf{S}}_{t+m}(x,y)] \geq \eta]$ 
\State $\mathbf{M}_{t+m+j}^{\text{final}} \gets \mathcal{C}_{t+m+j} \cdot \mathbf{M}_{t+m+j}$ 
\State \Return $\mathbf{M}_{t+m+j}^{\text{final}}$
\end{algorithmic}
\end{algorithm}

This entropy-aware vulnerability search strategy fundamentally differs from conventional greedy approaches by explicitly leveraging the information-theoretic vulnerability quantification established in Theorem 2. By maximizing entropy deficit $V(R) = H_c(R) - \frac{1}{n}\sum_{i=1}^{n}H_i(R)$ through the learned risk map $\tilde{\mathbf{S}}_{t+m}$, our approach systematically identifies optimal attack transformations across spatiotemporal dimensions that exploit regions where collective consensus uncertainty exceeds individual observation confidence—precisely the conditions under which CP defense mechanisms are most vulnerable to fabrication attacks.

After obtaining the final mask, we determine whether to perform full optimization or simply warp the perturbation. For frames where $\mathcal{C}_{t+m+j} = 1$ and we're within the persistence window $p$, we employ a feature-space homeomorphism $\mathcal{W}: \mathbb{R}^d \times \{0,1\}^{H \times W} \times \{0,1\}^{H \times W} \rightarrow \mathbb{R}^d$ that transfers the perturbation between frames:
\begin{equation}
F_{j}^{p} = \mathcal{W}(F_{j-1}^{p}, \mathbf{M}_{j-1}, \mathbf{M}_{j}),
\end{equation}
where $F_{j}^{p}$ represents the perturbed features at frame $t+m+j$, and $\mathbf{M}_{j}$ is shorthand for $\mathbf{M}_{t+m+j}^{\text{final}}$. This approach leverages the quasi-stationary nature of feature representations across consecutive frames, enabling efficient attack persistence without computationally expensive re-optimization.

\section{Occupancy Map Estimation}
\label{app:bsc_attack}

When occupancy maps are not explicitly shared in CP systems, we estimate them from available perception data by adapting the Blind Region Segmentation (BRS) algorithm proposed by Tao \textit{et al.} \citep{tao2025gcpguardedcollaborativeperception}.
The process begins with differential detection, where the attacker compares independent and collaborative perception results to infer each CAV's blind spots. For each collaborating CAV $j$, the attacker generates two perception outputs:
\begin{equation}
    \mathbf{Y}_{s}^{j} = \Phi_\mathtt{dec}(\mathbf{F}_{j \rightarrow i}), \quad \mathbf{Y}_{c}^{j} = \Phi_\mathtt{dec}(f_\mathtt{agg}(\mathbf{F}_{i \rightarrow i}, \mathbf{F}_{j \rightarrow i})),
\end{equation}
where $\mathbf{Y}_{s}^{j}$ represents single perception results using only CAV $j$'s features $\mathbf{F}_{j \rightarrow i}$, and $\mathbf{Y}_{c}^{j}$ represents collaborative perception results using both local features $\mathbf{F}_{i \rightarrow i}$ and CAV $j$'s features. The function $\Phi_\mathtt{dec}$ denotes the decoder network and $f_\mathtt{agg}$ is the feature aggregation function.
By computing the intersection $\mathbf{Y}_{s}^{j} \cap \mathbf{Y}_{c}^{j}$ and set difference operations, we identify:
\begin{equation}
    \mathbf{Y}_{j\text{-only}} = \mathbf{Y}_{s}^{j}, \quad \mathbf{Y}_{non\text{-}j} = \mathbf{Y}_{c}^{j} \setminus (\mathbf{Y}_{s}^{j} \cap \mathbf{Y}_{c}^{j}),
\end{equation}
where $\mathbf{Y}_{j\text{-only}}$ contains CAV $j$'s unique detections and $\mathbf{Y}_{non\text{-}j}$ contains detections not attributable to CAV $j$. This differential analysis reveals the spatial distribution of perception capabilities.
The BRS algorithm partitions the BEV detection map into confident areas (CA) and blind areas (BA) through an adaptive region growing process. The algorithm employs a spatially-aware neighbor selection function that adapts to the distance from CAV $j$'s center position $\mathbf{p}_j$:
\begin{equation}
    \kappa(\mathbf{s}, \mathbf{p}_j) = \left\lceil \kappa_0 \cdot \exp\left(-\lambda \cdot \frac{\|\mathbf{s} - \mathbf{p}_j\|_2}{\sqrt{H^2+W^2}}\right)\right\rceil,
\end{equation}
where $\kappa(\mathbf{s}, \mathbf{p}_j)$ determines the number of neighbors to consider at grid position $\mathbf{s}$, $\kappa_0=6$ is the base connectivity parameter, $\lambda=0.3$ controls the exponential decay rate, $\|\mathbf{s} - \mathbf{p}_j\|_2$ is the Euclidean distance between position $\mathbf{s}$ and CAV $j$'s center $\mathbf{p}_j$, and $\sqrt{H^2+W^2}$ normalizes the distance based on the BEV map dimensions $H \times W$. This adaptive connectivity ensures denser region growing near the CAV and sparser expansion in distant regions, modeling the natural degradation of perception reliability with distance.
The key step for our occupancy map estimation is converting the resulting binary mask $\mathbf{M}_{j} \in \{0,1\}^{H \times W}$ into an occupancy grid map $\mathbf{O}_j \in \{0,1,2\}^{H \times W}$. For each grid cell position $\mathbf{p} = (x,y)$, we apply the following mapping:
\begin{equation}
    \mathbf{O}_j(\mathbf{p}) = 
    \begin{cases}
        1, & \text{if } \mathbf{M}_{j}(\mathbf{p}) = 1 \text{ and } \mathbf{p} \in \mathcal{D}_j \\
        0, & \text{if } \mathbf{M}_{j}(\mathbf{p}) = 1 \text{ and } \mathbf{p} \notin \mathcal{D}_j \\
        2, & \text{if } \mathbf{M}_{j}(\mathbf{p}) = 0
    \end{cases}
\end{equation}
where $\mathcal{D}_j$ represents the set of grid positions containing detected objects from CAV $j$. In confident areas ($\mathbf{M}_{j}(\mathbf{p}) = 1$), grid cells containing detected objects are marked as occupied (1), while empty cells are marked as free (0). Blind areas ($\mathbf{M}_{j}(\mathbf{p}) = 0$) are marked as unknown (2). 
This process generates approximate occupancy grid maps for each collaborating CAV that, while not as precise as explicitly shared maps in CAD-enabled systems, provide sufficient information for constructing the MVIG and optimizing our attack strategy.

\begin{table*}[h]
    \centering
    \caption{MVIGNet Architecture and Parameters}
    \label{tab:mvig_structure}
    \renewcommand{\arraystretch}{1.2}
    \setlength{\tabcolsep}{10pt}
    \begin{tabular}{|>{\columncolor{lightgray!30}\centering\arraybackslash}m{2.8cm}|>{\ttfamily\small\centering\arraybackslash}m{4.2cm}|>{\centering\arraybackslash}m{5.7cm}|}
    \hline
    \rowcolor{lightgray!50}
    \multicolumn{1}{|>{\columncolor{lightgray!50}\centering\bfseries}c|}{Component} & 
    \multicolumn{1}{>{\columncolor{lightgray!50}\centering\normalfont\bfseries}c|}{Shape/Value} & 
    \multicolumn{1}{>{\columncolor{lightgray!50}\centering\bfseries}c|}{Description} \\
    \hline
    \multicolumn{3}{|>{\columncolor{lightgray!15}}c|}{\textbf{Input Data Structure}} \\
    \hline
    Temporal Graphs & List of Dictionaries & Sequence of graph structures \\
    \hline
    \multicolumn{3}{|>{\columncolor{lightgray!15}}c|}{\textbf{Node Features Composition}} \\
    \hline
    Basic Features & [3] & Normalized occupancy frequencies \\
    \hline
    Position-Pose & [6] & CAV position and orientation \\
    \hline
    Spatial Features & [91] & Multi-scale pooled features  \\
    \hline
    \multicolumn{3}{|>{\columncolor{lightgray!15}}c|}{\textbf{Edge Features Composition}} \\
    \hline
    Mutual Information & [1] & Average MI between occupancy maps \\
    \hline
    \multicolumn{3}{|>{\columncolor{lightgray!15}}c|}{\textbf{Backbone}} \\
    \hline
    Input & [batch, seq\_len, num\_nodes, 100] & Sequence of graph node features \\
    \hline
    MVIG-Conv ($\times 3$) & [batch, seq\_len, num\_nodes, 64] & Specialized graph convolution layers  \\
    \hline
    Mean Pooling & [batch, seq\_len, 64] & Aggregates node features per-frame \\
    \hline
    GRU & [batch, seq\_len, 64] & Processes temporal data across frames \\
    \hline
    Last Hidden & [batch, 64] & Extracts final hidden state from GRU \\
    \hline
    \multicolumn{3}{|>{\columncolor{lightgray!15}}c|}{\textbf{Score Head Component}} \\
    \hline
    FCN Layer 1 & [batch, 64] & Fully connected layer \\
    \hline
    ReLU & [batch, 64] & Activation function \\
    \hline
    FCN Layer 2 & [batch, 40000] & Fully connected layer \\
    \hline
    Softmax & [batch, 40000] & Converts to probability distribution \\
    \hline
    Grid Map & [batch, 200, 200] & Reshapes scores to 2D grid \\
    \hline
    Position Selection & [batch, 2] & Selects grid position \\
    \hline
    Output & [batch, 7] & bounding box parameters\\
    \hline
    \multicolumn{3}{|>{\columncolor{lightgray!15}}c|}{\textbf{Hyperparameters}} \\
    \hline
    node\_dim & 100 & Dimension of input node features \\
    \hline
    edge\_dim & 1 & Dimension of edge features \\
    \hline
    hidden\_dim & 64 & Dimension of hidden layers \\
    \hline
    num\_layers & 3 & Number of graph convolution layers \\
    \hline
    grid\_size & (200, 200) & Size of the output grid map \\
    \hline
    range\_limit & 20 & Range limit for x and y coordinates  \\
    \hline
    attack\_type & "spoof", "remove" & Type of attack to perform \\
    \hline
    \end{tabular}
\end{table*}

\begin{table*}[t]
    \centering
    \renewcommand{\arraystretch}{1.2}
    \setlength{\tabcolsep}{5.5pt}
    \small
    \caption{Comparison of existing attack methods in collaborative perception. \ding{51} indicates the feature is supported, while \ding{55} indicates it is not.}
    \label{tab:related_work}
    \begin{tabular}{l|c|c|c|c|cc|cc}
    \hline
    \multirow{2}{*}{\textbf{Method}} & \multirow{2}{*}{\textbf{System}} & \multirow{2}{*}{\textbf{Prerequisites}} & \multirow{2}{*}{\textbf{Real-time}} & \multirow{2}{*}{\textbf{Targeted}} & \multicolumn{2}{c|}{\textbf{Attack Type}} & \multicolumn{2}{c}{\textbf{Stealthiness}} \\
    \cline{6-9}
    & & & & & Spoof & Remove & Timing & Region \\
    \hline
    GPS/LiDAR spoofing \citep{9710897, 9519442} & Single & Laser emitters & \ding{51} & \ding{51} & \ding{51} & \ding{51} & \ding{55} & \ding{55} \\
    \rowcolor{gray!5} Physical attack \citep{9156447} & Single & Physical access & \ding{51} & \ding{51} & \ding{55} & \ding{51} & \ding{55} & \ding{55} \\
    Output manipulation \citep{9148235, s21030744} & Late fusion & None & \ding{51} & \ding{51} & \ding{51} & \ding{51} & \ding{55} & \ding{55} \\
    \rowcolor{gray!5} Basic feature attack \citep{9711249} & Int. fusion & Local computing & \ding{55} & \ding{55} & \ding{51} & \ding{55} & \ding{55} & \ding{55} \\
    RC attack \citep{294490} & Int. fusion & Local computing & \ding{51} & \ding{51} & \ding{51} & \ding{51} & \ding{55} & \ding{55} \\
    \rowcolor{gray!5} BAC attack \citep{tao2025gcpguardedcollaborativeperception} & Int. fusion & Local computing & \ding{55} & \ding{55} & \ding{51} & \ding{55} & \ding{55} & \ding{51} \\
    \rowcolor{yellow!10} \textbf{\texttt{MVIG} attack (Ours)} & Int. fusion & Local computing & \ding{51} & \ding{51} & \ding{51} & \ding{51} & \ding{51} & \ding{51} \\
    \hline
    \end{tabular}
  \end{table*}

\section{More Background and Related Work}
\label{app:background_related_work}

\subsection{Robust Collaborative Perception} 
Single-agent perception systems are constrained by limited field-of-view (FoV), which collaborative perception (CP) addresses through multi-agent data fusion \citep{hanCollaborativePerceptionAutonomous2023,huCollaborativePerceptionConnected2024}. CP architectures have evolved from raw-data-level fusion \citep{chen2023co} and output-level fusion \citep{10.1007/978-3-030-58589-1_10} to intermediate-level feature fusion, with DiscoNet \citep{NEURIPS2021_f702defb} and V2VNet \citep{10.1007/978-3-030-58536-5_36} demonstrating the effectiveness of this approach.
For practical deployment of CP systems, robustness against various challenges becomes crucial. These challenges can be broadly categorized into systematic robustness issues and malicious agent threats. Systematic robustness concerns include synchronization challenges \citep{lei2022latencyawarecollaborativeperception, NEURIPS2023_5a829e29}, communication interruption \citep{ren2024interruptionawarecooperativeperceptionv2x}, data corruption \citep{zhang2025dsrclearningdensityinsensitivesemanticaware}, and camera sensor failures \citep{NEURIPS2024_27e5626c}. While these system-level issues can affect perception performance, they are often predictable and can be mitigated through proper system design and error handling mechanisms.
In contrast, threats from malicious agents represent a more severe challenge to CP systems. Unlike systematic issues that occur randomly and independently, malicious agents can launch targeted attacks by deliberately manipulating shared information and exploiting system vulnerabilities. 

\subsection{Attacks on Collaborative Perception} 
The security challenges in CP systems have become increasingly complex due to the emergence of various adversarial attacks \citep{an2025removing, an2024box, yuan2025no, an2026decoder, an2025decoder}, which have evolved alongside advances in perception architectures. Initially, security concerns primarily focused on physical attacks against individual CAVs, including GPS spoofing \citep{9710897}, LiDAR spoofing \citep{9519442, 279980}, and the deployment of adversarial objects in the physical environment \citep{9156447}. As CP systems adopted collaborative frameworks, new vulnerabilities emerged. Late-fusion proved particularly susceptible to attacks, as their direct sharing of object detection results \citep{9148235, s21030744} made it straightforward for attackers to manipulate the perception outcomes \citep{9564757, 9575970}. Early attempts to attack these systems by corrupting raw sensor data proved ineffective against the more robust feature-level fusion systems that were later developed.

The evolution of CP systems towards feature-level fusion prompted attackers to develop more sophisticated strategies. Tu \textit{et al.} \citep{9711249} pioneered an untargeted adversarial attack framework by demonstrating how CP systems could be compromised through feature map perturbations, achieving high success rates. However, their approach produced obvious modifications of bounding boxes in perception results, making it easily detectable by threshold-based outlier anomaly detection methods \citep{Li_2023_ICCV,zhao2024maliciousagentdetectionrobust}. Tao \textit{et al.} \citep{tao2025gcpguardedcollaborativeperception} improved upon this by introducing the blind area confusion (BAC) attack that utilizes victim CAV's view information through differential detection and blind area estimation to generate a rough mask, reducing ineffective attack boxes while maintaining high success rates. Nevertheless, their approach relies solely on victim CAV's knowledge for mask estimation, failing to consider that benign agents can collectively validate detection results through their overlapping view regions.
Zhang \textit{et al.} \citep{294490} advanced this work by developing a targeted attack framework that uses specialized loss functions and masking techniques to generate fabricated detection boxes at specific locations. Their ray-casting (RC) attack achieved more stealthy and real-time attacks by leveraging LiDAR ray-casting to accelerate perturbation generation and enhance attack stealthiness. However, their method lacked systematic consideration of attack timing ("when to attack") and attack region selection ("where to attack"), relying instead on random selection of these critical parameters. This randomness made their attacks vulnerable to occupancy grid-based collaborative anomaly detection (CAD) systems. These limitations motivate our proposed \texttt{MVIG} attack, which systematically analyzes and exploits the dynamic mutual view relationships to achieve both effective and stealthy attacks through optimized attack timing and region control.
A comparison of the \texttt{MVIG} attack with existing attack methods is shown in Table \ref{tab:related_work}.

\subsection{Defensive Collaborative Perception} The growing security threats in CP systems have driven the development of various defense mechanisms. Initial defense approaches, including ROBOSAC \citep{Li_2023_ICCV}, CP-Guard \citep{hu2024cpguardmaliciousagentdetection}, and MADE \citep{zhao2024maliciousagentdetectionrobust}, primarily focused on detecting output-level perturbations through consensus-based verification, where perception outputs are cross-validated among CAVs using Hungarian matching and reconstruction loss. Building upon these foundations, Tao \textit{et al.} \citep{tao2025gcpguardedcollaborativeperception} enhanced the defense capabilities by introducing spatio-temporal anomaly detection to examine perception patterns across both spatial and temporal dimensions. While these methods demonstrated effectiveness against Tu's classical feature attack framework and naive attacks, they face significant challenges when confronted with sophisticated perturbations that maintain output-level consistency.

In response to more sophisticated attacks, Zhang \textit{et al.} \citep{294490} proposed a collaborative anomaly detection (CAD) system that implements a rule-based defense mechanism through occupancy map exchange. In their approach, the occupancy map is categorized into three states: 0 (free), 1 (occupied), and 2 (unknown). The defense operates on two principles: (1) in ego-known regions, any inconsistency with the ego CAV's occupancy map is immediately flagged as suspicious and evaluated against a conflict threshold to determine if an attack is present; (2) in unknown regions, the system relies on consistency checking among other collaborative CAVs' occupancy maps.
However, this defense strategy reveals critical vulnerabilities. The CAD system becomes ineffective in regions that are mutually unknown to benign CAVs or in scenarios where malicious agents have advantageous viewpoints. More importantly, the sharing of occupancy maps inadvertently provides attackers with valuable information about the collective perception coverage, allowing them to carefully optimize their attack timing and regions to bypass the defense mechanism.




\section{Structure of MVIGNet}
\label{app:mvig_structure}

The structure of the MVIGNet is shown in Table \ref{tab:mvig_structure}.

%% file: abbrev.bib
@STRING{jan = "Jan."}

@STRING{apr = "April"}

@STRING{may = "May"}

@STRING{jul = "July"}

@STRING{aug = "Aug."}

@STRING{oct = "Oct."}

@STRING{nov = "Nov."}


%% file: main.bib
@inproceedings {294490,
author = {Qingzhao Zhang and Shuowei Jin and Ruiyang Zhu and Jiachen Sun and Xumiao Zhang and Qi Alfred Chen and Z. Morley Mao},
title = {{On Data Fabrication in Collaborative Vehicular Perception: Attacks and Countermeasures}},
booktitle = {33rd USENIX Security Symposium (USENIX Security 24)},
year = {2024},
address = {Philadelphia, PA},
pages = {6309--6326},
month = aug
}

@INPROCEEDINGS{9711249,
  author={Tu, James and Wang, Tsunhsuan and Wang, Jingkang and Manivasagam, Sivabalan and Ren, Mengye and Urtasun, Raquel},
  booktitle={Proceedings of the IEEE/CVF International Conference on Computer Vision (ICCV)}, 
  title={{Adversarial Attacks On Multi-Agent Communication}}, 
  year={2021},
  volume={},
  number={},
  pages={7748-7757},
}

@InProceedings{Li_2023_ICCV,
    author    = {Li, Yiming and Fang, Qi and Bai, Jiamu and Chen, Siheng and Juefei-Xu, Felix and Feng, Chen},
    title     = {{Among Us: Adversarially Robust Collaborative Perception by Consensus}},
    booktitle = {Proceedings of the IEEE/CVF International Conference on Computer Vision (ICCV)},
    month     = {October},
    year      = {2023},
    pages     = {186-195}
}

@INPROCEEDINGS{zhao2024maliciousagentdetectionrobust,
  author={Zhao, Yangheng and Xiang, Zhen and Yin, Sheng and Pang, Xianghe and Wang, Yanfeng and Chen, Siheng},
  booktitle={IEEE/RSJ International Conference on Intelligent Robots and Systems (IROS)}, 
  title={{MADE}: {Malicious} {Agent} {Detection} for {Robust} {Multi}-{Agent} {Collaborative} {Perception}}, 
  year={2024},
  volume={},
  number={},
  pages={13817-13823},
}

@inproceedings{
  chen2023co,
  title={{CO}3: {Cooperative} {Unsupervised} 3D {Representation} {Learning} for {Autonomous} {Driving}},
  author={Runjian Chen and Yao Mu and Runsen Xu and Wenqi Shao and Chenhan Jiang and Hang Xu and Yu Qiao and Zhenguo Li and Ping Luo},
  booktitle={The Eleventh International Conference on Learning Representations (ICLR)},
  year={2023},
  }

@InProceedings{10.1007/978-3-030-58589-1_10,
author="Zeng, Wenyuan
and Wang, Shenlong
and Liao, Renjie
and Chen, Yun
and Yang, Bin
and Urtasun, Raquel",
editor="Vedaldi, Andrea
and Bischof, Horst
and Brox, Thomas
and Frahm, Jan-Michael",
title={{DSDNet}: {Deep} {Structured} {Self}-{Driving} {Network}},
booktitle="European Conference on Computer Vision (ECCV)",
year="2020",
pages="156--172",
}

@inproceedings{NEURIPS2021_f702defb,
 author = {Li, Yiming and Ren, Shunli and Wu, Pengxiang and Chen, Siheng and Feng, Chen and Zhang, Wenjun},
 booktitle = {Advances in Neural Information Processing Systems (NeurIPS)},
 editor = {M. Ranzato and A. Beygelzimer and Y. Dauphin and P.S. Liang and J. Wortman Vaughan},
 pages = {29541--29552},
 title = {{Learning Distilled Collaboration Graph for Multi-Agent Perception}},
 volume = {34},
 year = {2021}
}

@inproceedings{10.1007/978-3-030-58536-5_36,
author = {Wang, Tsun-Hsuan and Manivasagam, Sivabalan and Liang, Ming and Yang, Bin and Zeng, Wenyuan and Urtasun, Raquel},
title = {{V2VNet}: {Vehicle}-to-{Vehicle} {Communication} for {Joint} {Perception} and {Prediction}},
year = {2020},
booktitle = {European Conference on Computer Vision (ECCV)},
pages = {605-621},
numpages = {17},
location = {Glasgow, United Kingdom}
}

@inproceedings{NEURIPS2023_5a829e29,
 author = {Wei, Sizhe and Wei, Yuxi and Hu, Yue and Lu, Yifan and Zhong, Yiqi and Chen, Siheng and Zhang, Ya},
 booktitle = {Advances in Neural Information Processing Systems (NeurIPS)},
 editor = {A. Oh and T. Naumann and A. Globerson and K. Saenko and M. Hardt and S. Levine},
 pages = {28462--28477},
 title = {{Asynchrony-Robust Collaborative Perception via Bird's Eye View Flow}},
 volume = {36},
 year = {2023}
}

@inproceedings{Dosovitskiy17,
  title = {{CARLA}: {An} {Open} {Urban} {Driving} {Simulator}},
  author = {Alexey Dosovitskiy and German Ros and Felipe Codevilla and Antonio Lopez and Vladlen Koltun},
  booktitle = {Proceedings of the 1st Annual Conference on Robot Learning (CoRL)},
  pages = {1--16},
  year = {2017}
}

@InProceedings{Lang_2019_CVPR,
author = {Lang, Alex H. and Vora, Sourabh and Caesar, Holger and Zhou, Lubing and Yang, Jiong and Beijbom, Oscar},
title = {{PointPillars}: {Fast} {Encoders} for {Object} {Detection} {From} {Point} {Clouds}},
booktitle = {Proceedings of the IEEE/CVF Conference on Computer Vision and Pattern Recognition (CVPR)},
month = {June},
year = {2019}
}

@misc{tao2025gcpguardedcollaborativeperception,
      title={{GCP}: {Guarded} {Collaborative} {Perception} with {Spatial}-{Temporal} {Aware} {Malicious} {Agent} {Detection}}, 
      author={Yihang Tao and Senkang Hu and Yue Hu and Haonan An and Hangcheng Cao and Yuguang Fang},
      year={2025},
      eprint={2501.02450},
      archivePrefix={arXiv},
      primaryClass={cs.CV},
      note={arXiv:2501.02450 [cs]}
}

@article{hu2024cpguardmaliciousagentdetection, 
title={{CP-Guard}: Malicious Agent Detection and Defense in Collaborative Bird's Eye View Perception},
volume={39}, 
number={22}, 
journal={Proceedings of the AAAI Conference on Artificial Intelligence}, 
author={Hu, Senkang and Tao, Yihang and Xu, Guowen and Deng, Yiqin and Chen, Xianhao and Fang, Yuguang and Kwong, Sam}, 
year={2025}, 
month={Apr.}, 
pages={23203-23211} 
}

@INPROCEEDINGS{9710897,
  author={Li, Yiming and Wen, Congcong and Juefei-Xu, Felix and Feng, Chen},
  booktitle={IEEE/CVF International Conference on Computer Vision (ICCV)}, 
  title={{Fooling LiDAR Perception via Adversarial Trajectory Perturbation}}, 
  year={2021},
  volume={},
  number={},
  pages={7878-7887},
}

@INPROCEEDINGS{9519442,
  author={Cao, Yulong and Wang, Ningfei and Xiao, Chaowei and Yang, Dawei and Fang, Jin and Yang, Ruigang and Chen, Qi Alfred and Liu, Mingyan and Li, Bo},
  booktitle={IEEE Symposium on Security and Privacy (SP)}, 
  title={{Invisible for both Camera and LiDAR: Security of Multi-Sensor Fusion based Perception in Autonomous Driving Under Physical-World Attacks}}, 
  year={2021},
  volume={},
  number={},
  pages={176-194},
}

@inproceedings {279980,
author = {R. Spencer Hallyburton and Yupei Liu and Yulong Cao and Z. Morley Mao and Miroslav Pajic},
title = {{Security Analysis of {Camera-LiDAR} Fusion Against {Black-Box} Attacks on Autonomous Vehicles}},
booktitle = {31st USENIX Security Symposium (USENIX Security)},
year = {2022},
address = {Boston, MA},
pages = {1903--1920},
publisher = {USENIX Association},
month = aug
}

@INPROCEEDINGS{9156447,
  author={Tu, James and Ren, Mengye and Manivasagam, Sivabalan and Liang, Ming and Yang, Bin and Du, Richard and Cheng, Frank and Urtasun, Raquel},
  booktitle={IEEE/CVF Conference on Computer Vision and Pattern Recognition (CVPR)}, 
  title={{Physically Realizable Adversarial Examples for LiDAR Object Detection}}, 
  year={2020},
  volume={},
  number={},
  pages={13713-13722},
}

@INPROCEEDINGS{9148235,
  author={Hadded, Mohamed and Merdrignac, Pierre and Duhamel, Sacha and Shagdar, Oyunchimeg},
  booktitle={International Wireless Communications and Mobile Computing (IWCMC)}, 
  title={{Security attacks impact for collective perception based roadside assistance: A study of a highway on-ramp merging case}}, 
  year={2020},
  volume={},
  number={},
  pages={1284-1289},
}

@Article{s21030744,
AUTHOR = {Godoy, Jorge and Jiménez, Víctor and Artuñedo, Antonio and Villagra, Jorge},
TITLE = {{A Grid-Based Framework for Collective Perception in Autonomous Vehicles}},
JOURNAL = {Sensors},
VOLUME = {21},
YEAR = {2021},
NUMBER = {3},
ARTICLE-NUMBER = {744},
}

@INPROCEEDINGS{9564757,
  author={Boddupalli, Srivalli and Hegde, Ashwini and Ray, Sandip},
  booktitle={IEEE International Intelligent Transportation Systems Conference (ITSC)}, 
  title={{Replace}: Real-time Security Assurance in Vehicular Platoons Against V2V Attacks}, 
  year={2021},
  volume={},
  number={},
  pages={1179-1185},
}

@INPROCEEDINGS{9575970,
  author={Liu, Xiruo and Yang, Lily and Alvarez, Ignacio and Sivanesan, Kathiravetpillai and Merwaday, Arvind and Oboril, Fabian and Buerkle, Cornelius and Sastry, Manoj and Baltar, Leonardo Gomes},
  booktitle={IEEE Intelligent Vehicles Symposium (IV)}, 
  title={{MISO-V}: Misbehavior Detection for Collective Perception Services in Vehicular Communications}, 
  year={2021},
  volume={},
  number={},
  pages={369-376},
}

@inproceedings{
madry2018towards,
title={{Towards Deep Learning Models Resistant to Adversarial Attacks}},
author={Aleksander Madry and Aleksandar Makelov and Ludwig Schmidt and Dimitris Tsipras and Adrian Vladu},
booktitle={International Conference on Learning Representations (ICLR)},
year={2018},
}

@misc{lei2022latencyawarecollaborativeperception,
      title={{Latency-Aware Collaborative Perception}}, 
      author={Zixing Lei and Shunli Ren and Yue Hu and Wenjun Zhang and Siheng Chen},
      year={2022},
      eprint={2207.08560},
      archivePrefix={arXiv},
      primaryClass={cs.CV},
      note={arXiv:2207.08560 [cs]}
}

@misc{ren2024interruptionawarecooperativeperceptionv2x,
      title={{Interruption-Aware Cooperative Perception for V2X Communication-Aided Autonomous Driving}}, 
      author={Shunli Ren and Zixing Lei and Zi Wang and Mehrdad Dianati and Yafei Wang and Siheng Chen and Wenjun Zhang},
      year={2024},
      eprint={2304.11821},
      archivePrefix={arXiv},
      primaryClass={cs.RO},
      note={arXiv:2304.11821 [cs]}
}

@inproceedings{NEURIPS2024_27e5626c,
 author = {Wang, Tianhang and Lu, Fan and Zheng, Zehan and Chen, Guang and Jiang, Changjun},
 booktitle = {Advances in Neural Information Processing Systems (NeurIPS)},
 pages = {22350--22369},
 title = {{RCDN}: {Towards} {Robust} {Camera}-{Insensitivity} {Collaborative} {Perception} via {Dynamic} {Feature}-based 3D {Neural} {Modeling}},
 volume = {37},
 year = {2024}
}

@misc{zhang2025dsrclearningdensityinsensitivesemanticaware,
      title={{DSRC}: {Learning} {Density}-{Insensitive} and {Semantic}-{Aware} {Collaborative} {Representation} against {Corruptions}}, 
      author={Jingyu Zhang and Yilei Wang and Lang Qian and Peng Sun and Zengwen Li and Sudong Jiang and Maolin Liu and Liang Song},
      year={2025},
      eprint={2412.10739},
      archivePrefix={arXiv},
      primaryClass={cs.CV},
      note={arXiv:2412.10739 [cs]}
}

@inproceedings{chenCooperCooperativePerception2019,
	address = {Dallas, TX, USA},
	title = {{Cooper}: {Cooperative} {Perception} for {Connected} {Autonomous} {Vehicles} {Based} on {3D} {Point} {Clouds}},
	shorttitle = {Cooper},
	language = {en},
	urldate = {2024-05-30},
	booktitle = {IEEE 39th International Conference on Distributed Computing Systems (ICDCS)},
	publisher = {IEEE},
	author = {Chen, Qi and Tang, Sihai and Yang, Qing and Fu, Song},
	month = jul,
	year = {2019},
	pages = {514--524},
}

@article{liV2XSimMultiAgentCollaborative2022,
	title = {{V2X}-{Sim}: {Multi}-{Agent} {Collaborative} {Perception} {Dataset} and {Benchmark} for {Autonomous} {Driving}},
	volume = {7},
	shorttitle = {{V2X}-{Sim}},
	language = {en},
	number = {4},
	urldate = {2024-05-30},
	journal = {IEEE Robotics and Automation Letters (RA-L)},
	author = {Li, Yiming and Ma, Dekun and An, Ziyan and Wang, Zixun and Zhong, Yiqi and Chen, Siheng and Feng, Chen},
	month = oct,
	year = {2022},
	pages = {10914--10921},
}

@misc{huAdaptiveCommunicationsCollaborative2023,
	title = {{Adaptive Communications} in {Collaborative} {Perception} with {Domain} {Alignment} for {Autonomous} {Driving}},
	urldate = {2023-10-25},
	publisher = {arXiv},
	author = {Hu, Senkang and Fang, Zhengru and An, Haonan and Xu, Guowen and Zhou, Yuan and Chen, Xianhao and Fang, Yuguang},
	month = oct,
	year = {2023},
	note = {arXiv:2310.00013 [cs]},
}

@inproceedings{huWhere2commCommunicationefficientCollaborative2024,
	address = {Red Hook, NY, USA},
	title = {{Where2comm}: communication-efficient collaborative perception via spatial confidence maps},
	shorttitle = {Where2comm},
	urldate = {2024-06-03},
	booktitle = {Proceedings of the 36th {International} {Conference} on {Neural} {Information} {Processing} {Systems} ({NeurIPS})},
	author = {Hu, Yue and Fang, Shaoheng and Lei, Zixing and Zhong, Yiqi and Chen, Siheng},
	month = apr,
	year = {2024},
	pages = {4874--4886},
}

@inproceedings{xuOPV2VOpenBenchmark2022,
	address = {Philadelphia, PA, USA},
	title = {{OPV2V}: {An} {Open} {Benchmark} {Dataset} and {Fusion} {Pipeline} for {Perception} with {Vehicle}-to-{Vehicle} {Communication}},
	shorttitle = {{OPV2V}},
	language = {en},
	urldate = {2024-05-30},
	booktitle = {2022 {International} {Conference} on {Robotics} and {Automation} ({ICRA})},
	publisher = {IEEE},
	author = {Xu, Runsheng and Xiang, Hao and Xia, Xin and Han, Xu and Li, Jinlong and Ma, Jiaqi},
	month = may,
	year = {2022},
	pages = {2583--2589},
}

@article{hanCollaborativePerceptionAutonomous2023,
	title = {Collaborative {Perception} in {Autonomous} {Driving}: {Methods}, {Datasets} and {Challenges}},
	volume = {15},
	shorttitle = {Collaborative {Perception} in {Autonomous} {Driving}},
	urldate = {2023-11-20},
	number = {6},
	urldate = {2023-11-20},
	journal = {IEEE Intelligent Transportation Systems Magazine},
	author = {Han, Yushan and Zhang, Hui and Li, Huifang and Jin, Yi and Lang, Congyan and Li, Yidong},
	month = nov,
	year = {2023},
	note = {arXiv:2301.06262 [cs]},
	pages = {131--151},
}

@misc{huCollaborativePerceptionConnected2024,
	title = {Collaborative {Perception} for {Connected} and {Autonomous} {Driving}: {Challenges}, {Possible} {Solutions} and {Opportunities}},
	shorttitle = {Collaborative {Perception} for {Connected} and {Autonomous} {Driving}},
	urldate = {2024-01-04},
	publisher = {arXiv},
	author = {Hu, Senkang and Fang, Zhengru and Deng, Yiqin and Chen, Xianhao and Fang, Yuguang},
	month = jan,
	year = {2024},
	note = {arXiv:2401.01544 [cs, eess]},
}

@INPROCEEDINGS{tao_directcp,
  author={Tao, Yihang and Hu, Senkang and Fang, Zhengru and Fang, Yuguang},
  booktitle={2025 IEEE International Conference on Robotics and Automation (ICRA)}, 
  title={Directed-CP: Directed Collaborative Perception for Connected and Autonomous Vehicles via Proactive Attention}, 
  year={2025},
  pages={7004-7010}
}

@article{yuan2025no,
  sortkey={zzzzz},
  title={No Trespassing: Ground-view Adversarial Patches for Privacy-aware Management in COTS Robot Vacuum Cleaner},
  author={Yuan, Shuai and Xu, Guowen and Li, Hongwei and Zhang, Rui and Cao, Hangcheng and Qian, Xinyuan and Ni, Tao and Zhao, Qingchuan and Fang, Yuguang},
  journal={IEEE Transactions on Dependable and Secure Computing},
  year={2025},
  publisher={IEEE}
}

@article{an2026decoder,
  sortkey={zzzzz},
  title={Decoder Gradient Shields: A Family of Provable and High-Fidelity Methods Against Gradient-Based Box-Free Watermark Removal},
  author={An, Haonan and Hua, Guang and Du, Wei and Cao, Hangcheng and Tao, Yihang and Xu, Guowen and Rahardja, Susanto and Fang, Yuguang},
  journal={IEEE Transactions on Dependable and Secure Computing},
  year={2026},
  publisher={IEEE}
}

@inproceedings{an2025decoder,
  sortkey={zzzzz},
  title={Decoder Gradient Shield: Provable and High-Fidelity Prevention of Gradient-Based Box-Free Watermark Removal},
  author={An, Haonan and Hua, Guang and Fang, Zhengru and Xu, Guowen and Rahardja, Susanto and Fang, Yuguang},
  booktitle={Proceedings of the Computer Vision and Pattern Recognition Conference},
  pages={13424--13433},
  year={2025}
}

@inproceedings{an2025removing,
  sortkey={zzzzz},
  title={Removing Box-Free Watermarks for Image-to-Image Models via Query-Based Reverse Engineering},
  author={An, Haonan and Hua, Guang and Cao, Hangcheng and Fang, Zhengru and Xu, Guowen and Rahardja, Susanto and Fang, Yuguang},
  booktitle={The 40th Annual AAAI Conference on Artificial Intelligence},
  year={2025}
}

@article{an2024box,
  sortkey={zzzzz},
  title={Box-free model watermarks are prone to black-box removal attacks},
  author={An, Haonan and Hua, Guang and Lin, Zhiping and Fang, Yuguang},
  journal={arXiv preprint arXiv:2405.09863},
  year={2024}
}

@misc{guo2025neptunexactivextomaritimegeneration,
      sortkey={zzzzz},
      title={Neptune-X: Active X-to-Maritime Generation for Universal Maritime Object Detection}, 
      author={Yu Guo and Shengfeng He and Yuxu Lu and Haonan An and Yihang Tao and Huilin Zhu and Jingxian Liu and Yuguang Fang},
      year={2025},
      eprint={2509.20745},
      archivePrefix={arXiv},
      primaryClass={cs.CV}
}
